\documentclass[11pt]{article}
\pdfoutput=1
\usepackage{microtype}
\usepackage{graphicx}
\usepackage{subcaption}
\usepackage{booktabs} %
\usepackage{amssymb}
\usepackage{bm}
\usepackage{bbm}
\usepackage{amsmath}  %
\usepackage{amsthm}
\usepackage{xcolor}
\usepackage{textcomp}
\usepackage{multirow}
\usepackage{tabularx}
\usepackage{mathtools}
\usepackage[margin=1in]{geometry}
\usepackage{authblk}
\usepackage[numbers]{natbib}
\usepackage{xspace}
\usepackage{comment}
\usepackage{thmtools, thm-restate}  %
\usepackage{verbatim}

\usepackage{tikz}
\usetikzlibrary{arrows}
\usetikzlibrary{positioning}

\tikzset{
  treenode/.style = {align=center, inner sep=0pt, text centered,
    font=\sffamily},
  arn_n/.style = {treenode, circle, black, font=\sffamily\bfseries, draw=black,
    fill=white, text width=1.5em},%
  arn_r/.style = {treenode, circle, black, font=\sffamily\bfseries, draw=black,
    fill=white, text width=1.0em},%
  arn_x/.style = {treenode, rectangle, draw=black,
    minimum width=0.5em, minimum height=0.5em}%
}

\usepackage{hyperref}

\usepackage{bm}
\usepackage{enumitem}

\newcommand*{\ShowNotes}{}
\newcommand{\yell}[1]{\textcolor{black}{#1}}

\ifdefined\ShowNotes
  \newcommand{\colornote}[3]{{\color{#1}\bf{#2 #3}\normalfont}}
\else
  \newcommand{\colornote}[3]{}
\fi

\definecolor{darkred}{rgb}{0.7,0.1,0.1}
\definecolor{darkgreen}{rgb}{0.1,0.5,0.1}
\definecolor{violet}{rgb}{0.7,0.0,0.7}
\definecolor{dblue}{rgb}{0.2,0.2,0.8}
\definecolor{maroon}{rgb}{0.76,.13,.28}
\definecolor{burntorange}{rgb}{0.81,.33,0}
\definecolor{royalpurple}{rgb}{0.47,.31,0.66}

\ifdefined\ShowNotes
  
\else
  
\fi

\newcommand{\sysname}[0]{TABi}
\newcommand{\kbname}[0]{KILT-E}
\newcommand\ul[1]{\underline{#1}}

\newif\ifarxiv

\title{TABi: Type-Aware Bi-Encoders for Open-Domain Entity Retrieval}

\author{Megan Leszczynski, Daniel Y. Fu, Mayee F. Chen, and Christopher Ré 
\\  Department of Computer Science, Stanford University 
\\ \vspace{4mm} \{\href{mailto:mleszczy@cs.stanford.edu}{\tt mleszczy}, \href{mailto:danfu@cs.stanford.edu}{\tt danfu}, \href{mailto:mfchen@cs.stanford.edu}{\tt mfchen}, \href{mailto:chrismre@cs.stanford.edu}{\tt chrismre}\}\href{mailto:mleszczy@cs.stanford.edu}{\tt @cs.stanford.edu}
}

\date{}

\begin{document}

\maketitle

\begin{abstract}
  Entity retrieval---retrieving information about entity mentions in a query---is a key step in open-domain tasks, such as question answering or fact checking. 
However, state-of-the-art entity retrievers struggle to retrieve rare entities for ambiguous mentions due to biases towards popular entities.
Incorporating knowledge graph types during training could help overcome popularity biases, but there are several challenges: (1) existing type-based retrieval methods require mention boundaries as input, but open-domain tasks run on unstructured text, (2) type-based methods should not compromise overall performance, and (3) type-based methods should be robust to noisy and missing types.  
In this work, we introduce \sysname, a method to jointly train bi-encoders on knowledge graph types and unstructured text for entity retrieval for open-domain tasks. \sysname\ leverages a type-enforced contrastive loss to encourage entities and queries of similar types to be close in the embedding space. \sysname\ improves retrieval of rare entities on the Ambiguous Entity Retrieval (AmbER) sets, while maintaining strong overall retrieval performance on open-domain tasks in the KILT benchmark compared to state-of-the-art retrievers. 
\sysname\ is also robust to incomplete type systems, improving rare entity retrieval over baselines with only 5\% type coverage of the training dataset. 
We make our code publicly available.\footnote{\url{https://github.com/HazyResearch/tabi}}\footnote{Accepted to Findings of ACL 2022.}  
\end{abstract}

\section{Introduction}
Entity retrieval (ER) is the process of finding the most relevant entities in a knowledge base for a natural language query.\footnote{We use ER to refer to the page-level document retrieval setting, where entities correspond to Wikipedia pages.} ER is crucial for open-domain NLP tasks, where systems are provided with a query without the information needed to answer the query \cite{karpukhin-etal-2020-dense}. For instance, to answer the query, \emph{“What team does George Washington play for?”} an open-domain system can use an entity retriever to find information about George Washington in a knowledge base. Retrieving the correct George Washington in the query above---George Washington the baseball player, rather than George Washington the president---requires the retriever to recognize that keywords “team” and “play” imply George Washington is an athlete. However, recent work has shown that state-of-the-art retrievers exhibit popularity biases and struggle to resolve ambiguous mentions of rare ``tail" entities~\cite{chen-etal-2021-evaluating}.

\begin{figure*}
\centering
\includegraphics[width=6in]{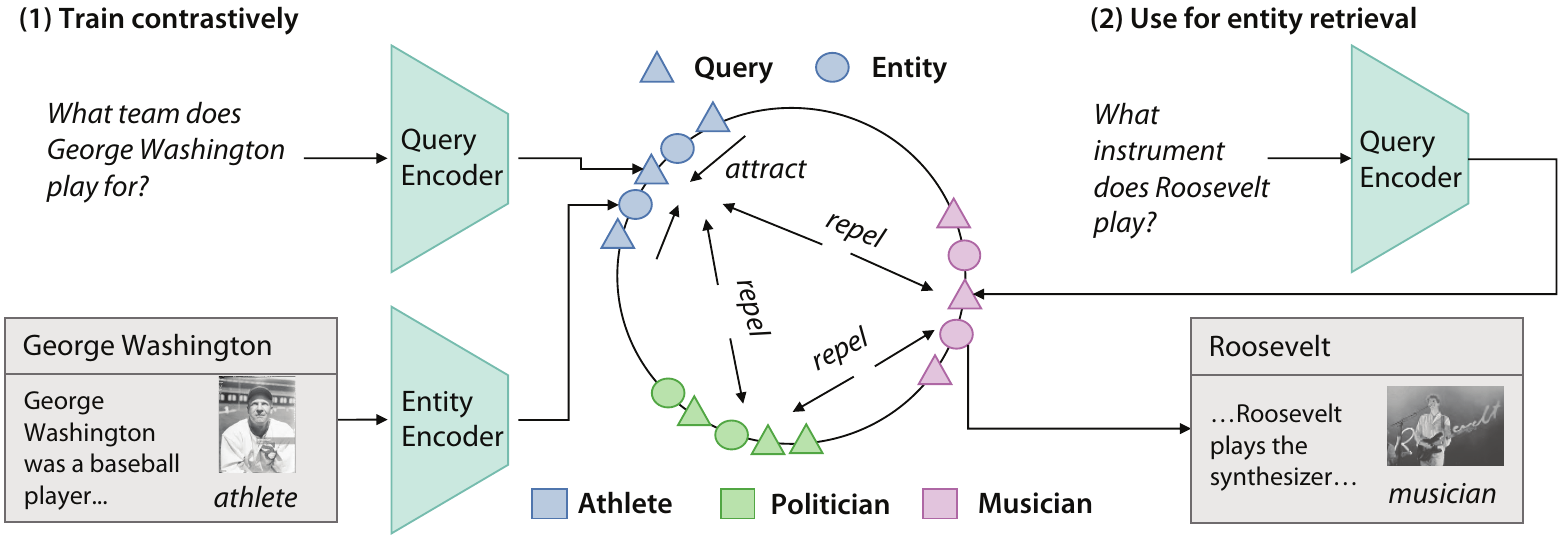}
\caption{\sysname\ uses a query and entity encoder to embed queries and entities in the same space. To encourage embeddings of the same type (e.g. athlete) to be close, \sysname\ introduces a type-enforced contrastive loss that pulls query embeddings of the same type together and pushes query embeddings of different types apart. 
}
\label{fig:banner}
\end{figure*}

The goal of our work is to improve rare entity retrieval for open-domain NLP tasks. Rare entities are challenging to retrieve when they share a name with more popular entities. For instance, in a sample of Wikipedia, mentions of George Washington refer to the president \yell{93\%} of the time, so a retriever can do very well by learning a popularity bias and returning the president whenever it sees “George Washington.” This strategy performs poorly on rare entities like George Washington the baseball player. To retrieve a rare entity instead of a more popular entity for an ambiguous mention, the retriever needs to learn to leverage context cues to overcome the popularity bias. However, existing state-of-the-art retrievers for open-domain tasks (e.g., GENRE~\cite{decao2020autoregressive}, DPR~\cite{karpukhin-etal-2020-dense}) are only trained on unstructured text, making it challenging for them to learn to associate context cues (e.g. “team” and “play”) with groups of entities (e.g., athletes).

A promising approach to overcome popularity biases is to incorporate types (e.g., athlete or politician) from a knowledge graph into the retriever. A key advantage of types is that contextual cues learned over popular entities can generalize to rare entities of the same types. However, there are several challenges with using types for open-domain retrieval. First, existing methods that use types assume mention boundaries are provided in the input~\cite{gupta-etal-2017-entity,onoe,bootleg}, but open-domain tasks run over unstructured text. These methods can suffer significant quality degradation without mention boundaries.\footnote{We find retrieval performance can drop 40\% (relative) by using mention detection v. gold mention boundaries.} Second, while it is important to do well on tail entities, the ideal retriever also needs to maintain strong performance over popular entities, balancing learning popularity biases with learning contextual cues. Finally, a retriever that incorporates types needs to be robust to incorrect and missing types, as type labels can be noisy and knowledge graphs can be incomplete.

In this work, we introduce \sysname, a method for training entity retrievers on knowledge graph types and unstructured text.
\sysname\ builds on the bi-encoder model for  dense retrieval (e.g., \citealp{wu-etal-2020-scalable,karpukhin-etal-2020-dense}) (Figure~\ref{fig:banner}). Bi-encoders learn embeddings of queries and entities contrastively: query embeddings are pulled close to their ground truth entity embedding and pushed away from other entity embeddings.

Our key insight is that type information should also be learned contrastively, as opposed to more straightforward approaches like adding the type as textual input.
\sysname\ adds a type-enforced contrastive loss term that pulls query embeddings of the same type together and pushes query embeddings of different types apart. As a result, \sysname\ clusters embeddings by type more strongly than simply adding the type as input or not using types at all (Figure~\ref{fig:tsne}), and thus performs better on nearest neighbor type classification and entity similarity tasks. Finally, motivated by “universal” dense retrievers~\cite{maillard-etal-2021-multi}, \sysname\ trains over multiple open-domain tasks in addition to entity disambiguation to support retrieval without mention boundaries.

Our experiments show that \sysname\ addresses the challenges of using types for open-domain retrieval. 
\textcolor{black}{First, we find that training a bi-encoder over multiple open-domain tasks significantly improves average top-1 tail retrieval by \yell{29.1} points compared to existing state-of-the-art baselines.
Our type-enforced loss further improves average top-1 tail retrieval by nearly 6 points.}
Second, \sysname\ maintains strong overall retrieval performance on popular entities, nearly matching or outperforming the state-of-the-art multi-task model, GENRE, on the eight open-domain KILT tasks ~\cite{kilt}. Third, \sysname\ is robust to missing and incorrect types, obtaining \yell{79\%} of the lift from the type-enforced loss even when only \yell{5\%} of the training examples have type annotations. Finally, we also explore a hybrid model that combines \sysname\ with a sparse retriever and popularity statistics. We find the hybrid model can lead to strong performance even when \sysname\ is trained without hard negative sampling, a standard but computationally expensive training procedure. To summarize, our contributions are as follows: 
\begin{itemize}[leftmargin=*,align=left]
    \item We introduce \sysname, a method to train bi-encoders on knowledge graph types and unstructured text through a new type-enforced contrastive loss for open-domain entity retrieval.
    \item We demonstrate that \sysname\ improves rare entity retrieval performance, maintains strong overall retrieval performance, and is robust to noisy and missing types on AmbER and KILT.  
    \item We validate that our approach can better capture types in query and entity embeddings than baseline dense entity retrievers through embedding visualization, nearest neighbor type classification, and an entity similarity task. 
\end{itemize}

\section{Preliminaries}
\label{sec:preliminaries} 

We review the problem setup, task, and the bi-encoder model.


\paragraph{Problem setup}
\label{ssec:prelim-setup}
Let $q \in \mathcal{Q}$ be a query, $e \in \mathcal{E}$ be an entity description, $y \in \mathcal{Y}$ be the entity label from the knowledge base, and $t \in \mathcal{T}$ be the type label.\footnote{To simplify notation, we define a single type label. In experiments, we define the type label as a set of entity types and type equivalence as 50\% of types matching (see Appendix~\ref{appendix:type-equivalence}).} 
We assume as input a labeled dataset $D = \{(q_{i}, e_{i}, y_i, t_i)\}^n_{i=1}$, where $n$ is the number of examples. 

\paragraph{Entity retrieval task}
\label{ssec:prelim-task}

Given a query $q$ as input, the entity retrieval task is to return the top-$K$ entity candidates relevant to the query from $\mathcal{Y}$. %
Since our primary motivation is open-domain NLP tasks, we focus on the page-level document retrieval setting, where we assume that each document corresponds to an entity (e.g., Wikipedia page) and that no mention boundaries are provided as input.  

\paragraph{Bi-encoders for entity retrieval}
\label{ssec:prelim-biencoders} 

The bi-encoder model consists of a query encoder $f: \mathcal{Q} \rightarrow \mathbb{R}^d$ and an entity encoder $g: \mathcal{E} \rightarrow \mathbb{R}^d$.
Most bi-encoders (e.g., \citealp{gillick-etal-2019-learning,wu-etal-2020-scalable}) are trained with the InfoNCE loss~\cite{Oord2018RepresentationLW}, in which ``positive'' pairs of examples are pulled together and ``negative'' pairs of examples are pushed apart. For a particular query $q$, let its positive example $e^+$ be the entity description for the respective gold entity and its negative examples $N_e(q)$ be the set of all other entity descriptions in the batch. 
For a batch with queries $Q$ and entity descriptions $E$, the loss is defined as: 
\begin{align}
L&_{NCE}(Q, E) = \frac{-1}{|Q|} \smashoperator{\sum_{q \in Q}} \nonumber \log \frac{\psi(q, e^+)}{\psi(q, e^+) + \smashoperator{\sum\limits_{e^- \in N_e(q)}}{ \psi(q, e^-)}},
\end{align}
where $\psi(v, w) = \exp(f(v)^\top g(w)/\tau)$ is the similarity score between the embeddings $v$ and $w$, and $\tau$ is a temperature hyperparameter. 
$L_{NCE}$ pulls each query embedding close to the entity embedding for its gold entity and pushes it away from all other entity embeddings in the batch. 
Batches are often constructed with hard negative samples to improve overall quality (e.g., \citealp{gillick-etal-2019-learning}).

\section{Approach} 
\label{sec:method}

\sysname\ leverages knowledge graph types and unstructured text to train bi-encoders for open-domain entity retrieval. 
\sysname\ takes as input queries and entity descriptions and uses a type-enforced contrastive loss.
At inference time, \sysname\ uses nearest neighbor search to retrieve entities.

\paragraph{Input} 
\label{ssec:input}
The query $q$ is represented as the WordPiece~\cite{Wu2016GooglesNM} tokens in the query, with special tokens \texttt{[M\textsubscript{s}]} and \texttt{[M\textsubscript{e}]} around the mention if the mention boundaries are known (matching the input of \citet{wu-etal-2020-scalable} with mention boundaries and \citet{karpukhin-etal-2020-dense} without). 
The entity description $e$ is represented as the first 128 WordPiece tokens of the entity's title and a description (i.e., Wikipedia page), with each component separated by an \texttt{[E\textsubscript{s}]} token, following \citet{wu-etal-2020-scalable}. 
We fine-tune the standard BERT-base pretrained model~\cite{devlin-etal-2019-bert} for both the query and entity encoders and take the final hidden layer representation corresponding to the \texttt{[CLS]} token as the query and entity embeddings. 
Similar to work in contrastive learning~\cite{simclr-chen}, we then apply L2 normalization to the embeddings.

\paragraph{Type-Enforced Contrastive Loss}
\label{ssec:loss}

We propose a contrastive loss that incorporates knowledge graph types and builds on the supervised contrastive loss from~\citet{supcon}.
Our goal is to encode types in the embedding space, such that the embeddings of queries and entities of the same type are closer together than those of different types. Types are often not sufficient to distinguish an entity, so we also want to embed queries and entities with similar names close together.

To achieve these two goals, our loss is a weighted sum of two supervised contrastive loss terms, $L_{type}$ and $L_{ent}$. 
For a randomly-sampled batch from dataset $D$ with queries $Q$ and entity descriptions $E$, \sysname's loss $L_{\sysname}$ is given by:
\begin{align}
    L_{\sysname}&(Q, E) = \alpha L_{type}(Q) + (1 - \alpha) L_{ent}(Q, E),
\end{align}
where $\alpha \in [0,1]$ (we use $\alpha = 0.1$ in our experiments).

$L_{type}(Q)$ uses type labels to form positive and negative pairs over queries.\footnote{We contrast queries in $L_{type}$ because we find it is more difficult to learn the query type than the entity type.}
Let $P_{type}(q)$ be the set of all queries in a batch that share the same type $t$ as a query $q$ and $N_{type}(q)$ be the other queries in the batch with a different type. 
Then $L_{type}(Q)$ is:
\begin{align}
    L&_{type}(Q) = \frac{-1}{|Q|} \sum_{q \in Q} \frac{1}{|P_{type}(q)|}~~~~\sum_{\mathclap{q^+ \in P_{type}(q)}} \;
    \log \frac{\psi(q, q^+)}{\psi(q, q^+) + \smashoperator{\sum\limits_{q^- \in N_{type}(q)}}{\psi(q, q^-)}}.
\end{align}

\noindent $L_{ent}(Q,E)$ uses entity labels to form positive and negative pairs over queries and entity descriptions.\footnote{In contrast, $L_{NCE}$ only compares query-entity pairs. We find that additionally comparing query-query and entity-entity pairs for $L_{ent}$ helps in \S \ref{ssec:ablations}.}
Let $x$ be a query or entity description, and $P_{ent}(x)$ be the set of all queries and entity descriptions in a batch that share the same gold entity $y$ as $x$.
Let $N_{ent}(x)$ be the set of all other queries and entity descriptions in the batch.
Then $L_{ent}(Q,E)$ is:
\begin{align}
    L&_{ent}(Q, E) = \frac{-1}{|Q \cup E|} \sum_{x \in Q \cup E}  \frac{1}{|P_{ent}(x)|}~~~~\sum_{\mathclap{x^+ \in P_{ent}(x)}} \;
    \log \frac{\psi(x, x^+)}{\psi(x, x^+) + \smashoperator{\sum\limits_{x^- \in N_{ent}(x)}}{\psi(x, x^-)}}.
\end{align}

We tie the weights of the query and entity encoders such that $f(\cdot) \equiv g(\cdot)$ so that $\psi$ is well-defined for all pairs of queries and entities.\footnote{Both encoders take a list of tokens as input.}
We also normalize embeddings before computing $\psi$. Following recent work~\cite{gillick-etal-2019-learning,karpukhin-etal-2020-dense}, we use hard negative sampling to add the top nearest incorrect entities for each query to the batch.\footnote{We train with three hard negatives for each query.} We follow \citet{botha-etal-2020-entity} to balance the hard negatives by fixing the ratio of positive to negative examples allowed for each entity, reducing the proportion of hard negatives that are rare entities (see Appendix~\ref{appendix:training-procedure}).

\begin{figure*}[t]
\center
\includegraphics[width=6in]{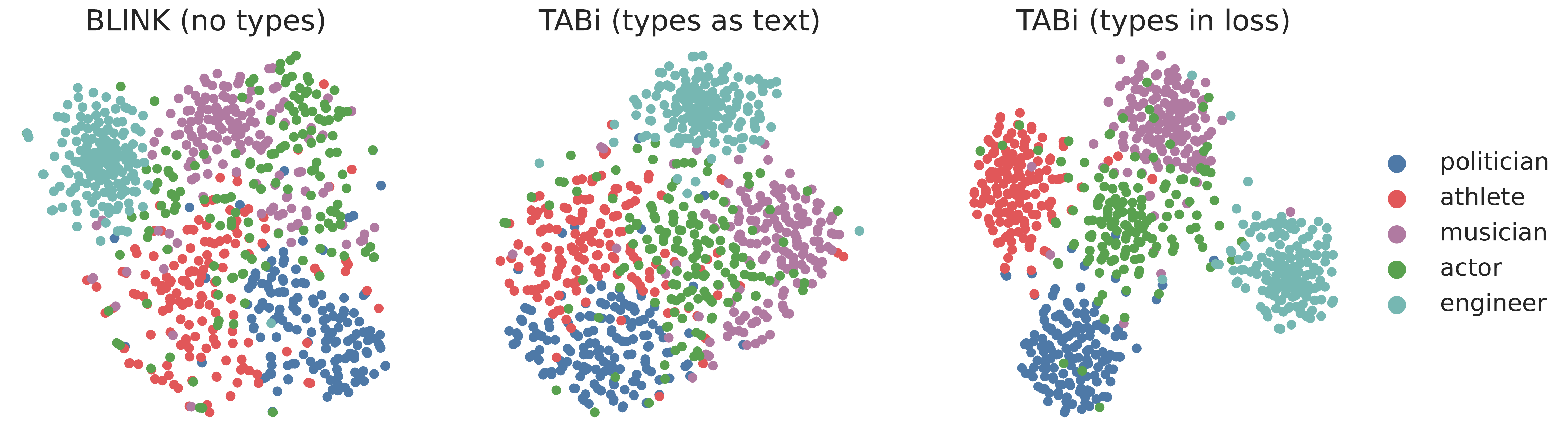}
\caption{t-SNE visualizations of entity embeddings. (Left) BLINK trained with $L_{NCE}$ without types. (Center) \sysname\ trained with $L_{ent}$ with types only as text in the input. (Right) \sysname\ trained with the type-enforced loss $L_{TABi}$.}
\label{fig:tsne}
\end{figure*}

The key difference between $L_{type}$ and $L_{ent}$ is the set of positive and negative pairs. 
$L_{type}$ forms pairs by type, which clusters queries of the same type in the embedding space. 
$L_{ent}$ forms pairs by gold entity, which clusters queries and entities with similar names in the embedding space. 
Figure~\ref{fig:tsne} shows that $L_{\sysname}$ produces embeddings that cluster better by types than those produced by $L_{NCE}$ (BLINK~\cite{wu-etal-2020-scalable}) or $L_{ent}$ with types simply added as text to the entity encoder input.

\paragraph{Inference} 
\label{ssec:inference} 
We precompute entity embeddings and use nearest neighbor search to retrieve the top-$K$ most similar entity embeddings to a query embedding.
While our standard configuration does not use a re-ranker, in Section~\ref{ssec:rerank} we also study the impact of adding an inexpensive re-ranker which linearly combines \sysname's scores with sparse retriever scores and popularity statistics (see Appendix~\ref{appendix:rerank}). 
Prior work has shown that a hybrid model that combines sparse retrievers (e.g. TF-IDF) and dense retrievers can improve performance~\cite{karpukhin-etal-2020-dense,Luan2021SparseDA} and that entity popularity can help disambiguation~\cite{ganea-hofmann-2017-deep}.

\section{Retrieval Experiments} 
\label{sec:experiments} 
Our experiments find that \sysname\ can improve rare entity retrieval for open-domain NLP tasks while maintaining strong overall retrieval performance.

\subsection{Experimental setup}
\label{ssec:experimental-setup}
We describe the baselines, evaluation datasets, knowledge base, and training data. We include additional setup details in Appendix~\ref{appendix:experimental-setup}. 

\paragraph{Baselines}

We compare against text-only baselines, which do not use types, to evaluate to what extent using types can improve performance over existing methods. We also compare against type-aware baselines, which use types and text, to better understand the challenges with incorporating types.

\begin{itemize}[leftmargin=*,align=left,topsep=1ex,itemsep=0ex,partopsep=0ex,parsep=1ex]
    \item \emph{Text-only baselines:} Alias Table sorts candidates by their prior probabilities with the mention in the BLINK training dataset. TF-IDF uses sparse embeddings of normalized word frequencies. DPR~\cite{karpukhin-etal-2020-dense} is a dense passage retriever that does not use mention boundaries. BLINK (Bi-encoder)~\cite{wu-etal-2020-scalable} is a state-of-the-art dense entity retriever which uses mention boundaries; we also compare against BLINK with a cross-encoder to re-rank the top 10 candidates from the bi-encoder. ELQ~\cite{li-etal-2020-efficient} finetunes the BLINK bi-encoder jointly with mention detection and entity disambiguation tasks. GENRE~\cite{decao2020autoregressive} is an autoregressive retriever that generates the full entity name from the mention. We use pretrained models for all text-only baselines, with the exception of Alias Table and TF-IDF, which are non-learned. 
    \item \emph{Type-aware baselines:} Bootleg~\cite{bootleg} is a Transformer-based model that re-ranks candidates from an alias table using types and knowledge graph relations. We also introduce two baselines for encoding types in open-domain retrievers: GENRE-type and \sysname-type-text. GENRE-type includes the types as part of the entity name, and thus must generate the entity name along with its types. \sysname-type-text adds the types as textual input to the entity encoder instead of the loss function and uses $L_{ent}$ for training. We use a pretrained model for Bootleg, fine-tune a pretrained model of GENRE to create GENRE-type, and fine-tune \sysname-type-text from a BERT-base pretrained model~\cite{devlin-etal-2019-bert}.   
\end{itemize}

\paragraph{Evaluation datasets} 
We use 14 datasets from two benchmarks: Ambiguous Entity Retrieval (AmbER)~\cite{chen-etal-2021-evaluating} and Knowledge Intensive Language Tasks (KILT)~\cite{kilt}. 
AmbER evaluates retrieval of ambiguous rare entities, and KILT evaluates overall retrieval performance.

\emph{AmbER.} AmbER~\cite{chen-etal-2021-evaluating} spans three tasks in open-domain NLP---fact checking, slot filling, and question answering---and is divided into human and non-human subsets, for a total of 6 datasets. AmbER tests the ability to retrieve the correct entity when at least two entities share a name (i.e. are ambiguous). The queries are designed to be resolvable, such that each query should contain enough information to retrieve the correct entity. AmbER also comes with ``head" (i.e. popular) and ``tail" (i.e. rare) labels, using Wikipedia page views for popularity. We split AmbER into dev and test (5/95 split) and report on the test set.\footnote{We use AmbER dev to select re-ranker hyperparameters in Section~\ref{ssec:rerank}.}

We create a variant of this dataset--AmbER (GOLD)--with gold mention boundaries.
While we focus on open-domain tasks, where mention boundaries are often unknown, AmbER (GOLD) enables us to evaluate disambiguation in isolation. 

Following \citet{chen-etal-2021-evaluating}, we report accuracy@1 (i.e. top-1 retrieval accuracy), which is the percentage of queries where the top-ranked entity is the gold entity. 
As multiple entities share a name with the query mention (by the dataset definition), this metric captures how well a model can use context to disambiguate.

\emph{KILT.} We consider 8 evaluation datasets across the four open-domain tasks in the KILT~\cite{kilt} benchmark (fact checking (FC), question answering (QA), slot filling (SF), and dialogue). All examples have been annotated with the Wikipedia page(s) that help complete the task. 

Following \citet{kilt}, we report R-precision~\cite{Beitzel2009}. Given $R$ gold entities, R-precision is equivalent to the proportion of relevant entities in the top-R ranked entities. With the exception of FEVER and HotPotQA, which may require multiple entities, R-precision is equivalent to accuracy@1. We compare against published and leaderboard numbers for KILT and refer the reader to \citet{kilt} for baseline details.

\paragraph{Knowledge base} We create a filtered version of the KILT knowledge base~\cite{kilt} with \yell{5.45M} entities that correspond to English Wikipedia pages. We remove Wikimedia internal items (e.g., disambiguation pages, list articles) from the KILT knowledge base, since they do not refer to real-world entities. We refer to our knowledge base as \kbname\ (KILT-Entity) and use it for all models at inference time for fair comparison.\footnote{As an exception, we report existing numbers for baselines with the full KILT knowledge base (5.9M entities) on the KILT benchmark test sets due to a benchmark submission limit. See Appendix~\ref{appendix:extended_kilt} for dev results with \kbname\ knowledge base.}

\paragraph{Training data} We train two versions of \sysname\ to understand the performance with and without mention boundaries in the input. 
For retrieval experiments with mention boundaries and embedding quality experiments, we train on the BLINK~\cite{wu-etal-2020-scalable} training data, which consists of 8.9M Wikipedia sentences.\footnote{We remove examples with gold entities not in KILT-E.}
For retrieval experiments without mention boundaries, we follow \citet{decao2020autoregressive} and train on all KILT training data (which includes open-domain tasks) and contains \yell{11.7M} sentences~\cite{kilt}.  
For type labels, we use the 113 types from the FIGER~\cite{Ling2012FineGrainedER} type set. 
To assign entity types, we use a direct mapping of Wikidata entities to Freebase entities to find the FIGER types associated with each entity in Freebase. To assign query types, we follow \citet{Ling2012FineGrainedER} and add the types of the gold entity for each query as the query type labels. 
While types can be incomplete and not present in the query, we find that the type labels are sufficient for improving the embedding quality (\S \ref{sec:type-eval}).

\subsection{Results}
\label{ssec:results}

\begin{table*}[t]
\centering
\fontsize{8.4}{10.1}\selectfont 
\setlength{\tabcolsep}{4.4pt}
\begin{tabularx}{\textwidth}{l rr rr rr rr rr rr rr}
\toprule
           & \multicolumn{4}{c}{\bf{Fact Checking}}   & \multicolumn{4}{c}{\bf{Slot Filling}}        & \multicolumn{4}{c}{\bf{Question Answering}} &    &      \\
           & \multicolumn{2}{c}{H}     & \multicolumn{2}{c}{N}   & \multicolumn{2}{c}{H}     & \multicolumn{2}{c}{N} & \multicolumn{2}{c}{H}     & \multicolumn{2}{c}{N} & \multicolumn{2}{c}{\bf{Average}} \\
           \cmidrule(lr){2-3}
           \cmidrule(lr){4-5}
           \cmidrule(lr){6-7}
           \cmidrule(lr){8-9}
           \cmidrule(lr){10-11}
           \cmidrule(lr){12-13}
           \cmidrule(lr){14-15}
           \bf{Model} & Head & Tail & Head & Tail & Head & Tail & Head & Tail & Head               & Tail & Head & Tail & Head    & Tail \\
\midrule      
TF-IDF                     & 27.8      & 29.3      & 23.0       & 21.8      & 26.7      & 23.5       & 17.3       & 13.7        & 24.2      & 22.6      & 18.2      & 13.9      & 22.9      & 20.8      \\
DPR                        & 25.3      & 14.3      & \ul{47.7}   & 23.7      & 13.9      & 5.1        & 48.6       & 22.2        & 21.0      & 8.8       & 52.1      & 23.4      & 34.8      & 16.3      \\
BLINK (Bi-encoder)         & 56.4      & 52.0      & 24.8       & 10.5      & \bf{76.8} & 55.7       & 30.7       & 13.5        & \ul{78.3}      & 55.7      & 67.3      & 33.8      & 55.7      & 36.9      \\
BLINK                     & 55.8      & 45.8      & 7.4        & 3.9       & 74.7      & 30.3       & 32.1       & 16.1        & \bf{83.8} & 43.8      & 71.3      & 44.5      & 54.2      & 30.7      \\
ELQ                        & 43.5      & 37.4      & 5.3        & 2.2       & 74.4      & 44.1       & 59.5       & 27.1        & 77.5      & 47.2      & 62.0      & 30.7      & 53.7      & 31.4      \\
GENRE                      & 59.9      & 30.7      & 32.6       & 19.9      & 67.1      & 52.6       & 72.9       & 59.5        & 62.9      & 28.4      & 61.1      & 32.4      & 59.4      & 37.2      \\
\midrule
Bootleg$^\dagger$           & 48.7      & 37.0      & 3.7        & 2.5       & 65.1      & 48.0       & 47.5       & 26.7        & 74.8      & 48.0      & 60.5      & 44.2      & 50.0      & 34.4      \\
GENRE-type & 32.2 & 50.6 & \bf{55.7} & 34.9 & 34.9 & 68.0 & 75.4 & 69.6 & 41.6 & 55.8 & 72.1 & 47.6 & 52.0 & 54.4 \\
\sysname-type-text & \ul{76.7} & \ul{60.4} & 39.0 & \ul{36.8} & 71.6 & \ul{86.3} & \ul{82.5} & \ul{85.2} & 69.6 & \ul{66.1} & \ul{82.3} & \ul{57.0} & \ul{70.3} & \ul{65.3}  \\
\bf{\sysname}  & \bf{83.5} & \bf{73.3} & 40.7 & \bf{41.7} & \ul{75.1} & \bf{89.4} & \bf{85.6} & \bf{88.0} & 78.0 & \bf{74.3} & \bf{83.0} & \bf{66.1} & \bf{74.3} & \bf{72.1} \\
\midrule
\sysname\ ($\alpha=0$) & 77.6 & 61.9 & 41.4 & 39.1 & 70.9 & 87.1 & 83.2 & 85.9 & 72.5 & 66.3 & 82.2 & 57.7 & 71.3 & 66.3 \\
\sysname\ ($L_{type}$+$L_{NCE}$) & 80.5 & 64.7 & 42.0 & 42.2 & 69.1 & 87.7 & 83.9 & 87.3 & 72.2 & 67.7 & 81.3 & 61.8 & 71.5 & 68.6 \\
\bottomrule
\end{tabularx}
\caption{Retrieval accuracy@1 on AmbER (H for human, N for non-human subsets). (Top) text-only methods, (middle) type-aware methods, and (bottom) ablations. $^\dagger$Models with an alias table. See Section~\ref{ssec:results} for training data details. Best score \textbf{bolded}, second best \ul{underlined} (excluding ablations).}
\label{tab:amber-results}
\end{table*}

\begin{table*}[t]
\bigskip
    \centering
    \fontsize{8.4}{10.1}\selectfont 
    \setlength{\tabcolsep}{4.5pt}
    \begin{tabularx}{\textwidth}{l rr rr rr rr rr rr rr}
    \toprule
               & \multicolumn{4}{c}{\bf{Fact Checking}}   & \multicolumn{4}{c}{\bf{Slot Filling}}        & \multicolumn{4}{c}{\bf{Question Answering}} &    &      \\
               & \multicolumn{2}{c}{H}     & \multicolumn{2}{c}{N}   & \multicolumn{2}{c}{H}     & \multicolumn{2}{c}{N} & \multicolumn{2}{c}{H}     & \multicolumn{2}{c}{N} & \multicolumn{2}{c}{\bf{Average}} \\
               \cmidrule(lr){2-3}
               \cmidrule(lr){4-5}
               \cmidrule(lr){6-7}
               \cmidrule(lr){8-9}
               \cmidrule(lr){10-11}
               \cmidrule(lr){12-13}
               \cmidrule(lr){14-15}
               \bf{Model} & Head & Tail & Head & Tail & Head & Tail & Head & Tail & Head               & Tail & Head & Tail & Head    & Tail \\
    \midrule        
    Alias Table$^\dagger$  & 45.9       & 6.6       & 45.8      & 7.9          & 45.9      & 6.5        & 45.7      & 7.8       & 45.7       & 6.5       & 45.3      & 7.9       & 45.7      & 7.2  \\
    TF-IDF                     & 27.8       & 29.3      & 23.0      & 21.8         & 26.7      & 23.5       & 17.3      & 13.7      & 24.2       & 22.6      & 18.2      & 13.9      & 22.9      & 20.8 \\
    BLINK (Bi-encoder)         & 77.5       & 66.5      & 77.0      & 46.0         & 76.9 & 55.9       & 63.8      & 29.9      & 78.4       & 55.8      & 71.0      & 34.8      & 74.1      & 48.2 \\
    BLINK                     & 81.8  & 61.0      & \ul{81.6} & \ul{58.5}         & 75.4      & 30.5       & 64.8 & 35.7      & \ul{83.8}  & 43.9      & \ul{74.9} & 45.7      & \ul{77.1} & 45.9 \\
        GENRE                      & 70.9       & 44.5      & 72.9      & 40.6         & 70.6      & 39.0       & 64.8 & 33.1      & 71.1       & 40.6      & 70.3      & 40.0      & 70.1      & 39.6 \\
        \midrule
    Bootleg$^\dagger$          & \ul{83.0}  & 70.7      & \bf{82.1} & 56.6         & \bf{84.9} & 58.8       & \bf{76.1}      & \bf{54.7} & \bf{86.3}  & 51.2      & \bf{79.2} & \bf{56.5} & \bf{82.0} & \ul{58.1} \\
    GENRE-type & 69.7 & 60.8 & 75.9 & 48.5 & 70.9 & 54.3 & \ul{66.7} & 37.2 & 70.7 & 54.6 & 72.5 & 46.7 & 71.1 & 50.3
 \\
 \sysname-type-text & 81.5 & \ul{75.0} & 78.9 & 58.1 & \ul{78.5} & \ul{62.1} & 63.1 & 38.6 & 80.0 & \ul{61.5} & 68.2 & 42.0 & 75.0 & 56.2
 \\
  \bf{\sysname} & \bf{84.4} & \bf{82.3} & 80.4 & \bf{63.5} & \ul{78.5} & \bf{68.6} & 64.5 & \ul{39.1} & 81.5 & \bf{69.8} & 71.8 & \ul{51.6} & 76.9 & \bf{62.5} \\
    \bottomrule
    \end{tabularx}
    \caption{Retrieval accuracy@1 on AmbER (GOLD) (with mention boundaries). (Top) text-only, (bottom) type-aware methods. All models are trained on Wikipedia. $^\dagger$Models with an alias table. Best score \textbf{bolded}, second best \ul{underlined}. 
    }
    \label{tab:amber-gold-results}
\end{table*}

\paragraph{Rare entities} \sysname\ improves retrieval of rare entities for ambiguous mentions.
On AmbER, \sysname\ improves average tail accuracy@1 by \yell{34.9} points compared to existing text-only baselines and \yell{6.8} points compared to type-aware baselines (Table~\ref{tab:amber-results}). 
Note that GENRE, GENRE-type, \sysname-type-text, and \sysname\ are trained on KILT data (which includes open-domain tasks), while BLINK, ELQ, and Bootleg are trained on Wikipedia entity disambiguation data, and DPR is trained on question answering data. 
See the ablations for a discussion of the training data impact. 
On AmbER (GOLD) where all models are trained on Wikipedia entity disambiguation data and mention boundaries are available (Table~\ref{tab:amber-gold-results}),
\sysname\ outperforms baselines on average tail accuracy@1 by \yell{4.4} points. 
BLINK and Bootleg perform much better on AmbER (GOLD) than on AmbER, suggesting that mention detection introduces significant error.

\paragraph{Overall performance} 
\sysname\ maintains strong performance overall.
On AmbER, \sysname\ outperforms all retrievers for average accuracy@1 over the head (Table~\ref{tab:amber-results}). 
On AmbER (GOLD), \sysname\ follows
Bootleg, which leverages an alias table limiting the number of candidates, and BLINK, which uses an expensive cross-encoder for re-ranking (Table~\ref{tab:amber-gold-results}). 
On KILT, we find that \sysname\ outperforms GENRE, the best performing multi-task retriever\footnote{\sysname\ and GENRE use a single model across all tasks, whereas KGI~\cite{glass-etal-2021-robust} and Re2G (anonymous), train a separate model for each task.} overall by \yell{1 point} and sets the state-of-the-art on three KILT tasks  (Table~\ref{tab:kilt-results}).

\begin{table*}[t]
\centering
\fontsize{8.4}{10.1}\selectfont 
\setlength{\tabcolsep}{9pt}
\begin{tabular}{lccccccccc}
\toprule
 &\multicolumn{1}{c}{\bf{Fact Check.}}   & \multicolumn{2}{c}{\bf{Slot Filling}}        & \multicolumn{4}{c}{\bf{Question Answering}} &  \multicolumn{1}{c}{\bf{Dial.}}       \\
 \cmidrule(lr){2-2} \cmidrule(lr){3-4} \cmidrule(lr){5-8} \cmidrule(lr){9-9}
 & \multicolumn{1}{c}{FEV}  & \multicolumn{1}{c}{T-REx} & \multicolumn{1}{c}{zsRE}   & \multicolumn{1}{c}{NQ} & \multicolumn{1}{c}{HoPo}  & \multicolumn{1}{c}{TQA}  & \multicolumn{1}{c}{ELI5}  & \multicolumn{1}{c}{WoW} & \bf{Avg.}\\
\midrule
TF-IDF*         & 50.9 & 44.7 & 60.8 & 28.1 & 34.1 & 46.4 & 13.7 & 49.0 & 41.0 \\
DPR*            & 55.3 & 13.3 & 28.9 & 54.3 & 25.0 & 44.5 & 10.7 & 25.5 & 32.2 \\
Multi-task DPR* & 74.5 & 69.5 & 80.9 & 59.4 & 42.9 & 61.5 & 15.5 & 41.1 & 55.7 \\
BLINK*    & 63.7 & 59.6 & 78.8 & 24.5 & 46.1 & 65.6 & 9.3  & 38.2 & 48.2 \\
GENRE$^\dagger$  & 83.6 & 79.4 & 95.8 & 60.3 & \ul{51.3} & 69.2 & \ul{15.8} & \bf{62.9} &  \ul{64.8} \\
KGI**  & 75.6 & 74.4 & \bf{98.5} & \ul{63.7} &\multicolumn{1}{c}{-} & 60.5 & \multicolumn{1}{c}{-} & 55.4 & \multicolumn{1}{c}{-}  \\
Re2G** & \bf{88.9} & \ul{80.7} & \multicolumn{1}{c}{-}  & \bf{70.8} & \multicolumn{1}{c}{-} & \bf{72.7} & \multicolumn{1}{c}{-} & \ul{60.1} & \multicolumn{1}{c}{-} \\ 
\bf{\sysname} &  \ul{84.4} & \bf{81.9} & \ul{96.2} & 62.6 & \bf{53.1} & \ul{70.4} & \bf{18.3} & 59.1 & \bf{65.8} \\
\bottomrule
\end{tabular}
\caption{R-precision on KILT open-domain tasks (test data). *Numbers from \citet{kilt}. $^\dagger$Numbers from \citet{decao2020autoregressive}. **Numbers from KILT leaderboard. Best score \textbf{bolded} and second best \ul{underlined}.}
\label{tab:kilt-results}
\end{table*}

\paragraph{Ablations} 
\label{ssec:ablations}
\textcolor{black}{
Table~\ref{tab:amber-results} reports ablations. 
First, to measure the impact of types, we remove the type-enforced loss ($L_{type}$) by setting $\alpha=0$. 
This is equivalent to training \sysname\ with just $L_{ent}$. 
Compared to the standard \sysname\ setting, the average accuracy@1 drops by \yell{5.8} and \yell{3.0} points on the tail and head, demonstrating the importance of the type-enforced loss, particularly over the tail. 
Moreover, we observe that \sysname\ ($\alpha=0$) still outperforms the BLINK bi-encoder by \yell{29.4} points over the tail. 
As the BLINK bi-encoder is trained only on entity disambiguation, this suggests that additionally training over open-domain tasks leads to substantial improvements (see Appendix~\ref{appendix:extended_amber}).} 
Second, we evaluate the impact of using $L_{ent}$ instead of the standard $L_{NCE}$ to compare pairs of queries and entity descriptions based on their gold entity (Section~\ref{ssec:loss}). Compared to the standard \sysname\ setting (which uses $L_{type}+L_{ent}$), \sysname\ ($L_{type}+L_{NCE}$) incurs an average accuracy@1 drop of \yell{3.5} and \yell{2.8} points over the tail and head, respectively.

\begin{figure*}[t]
\centering
\begin{subfigure}[t]{0.33\textwidth}
\centering
\includegraphics[height=1.2in]{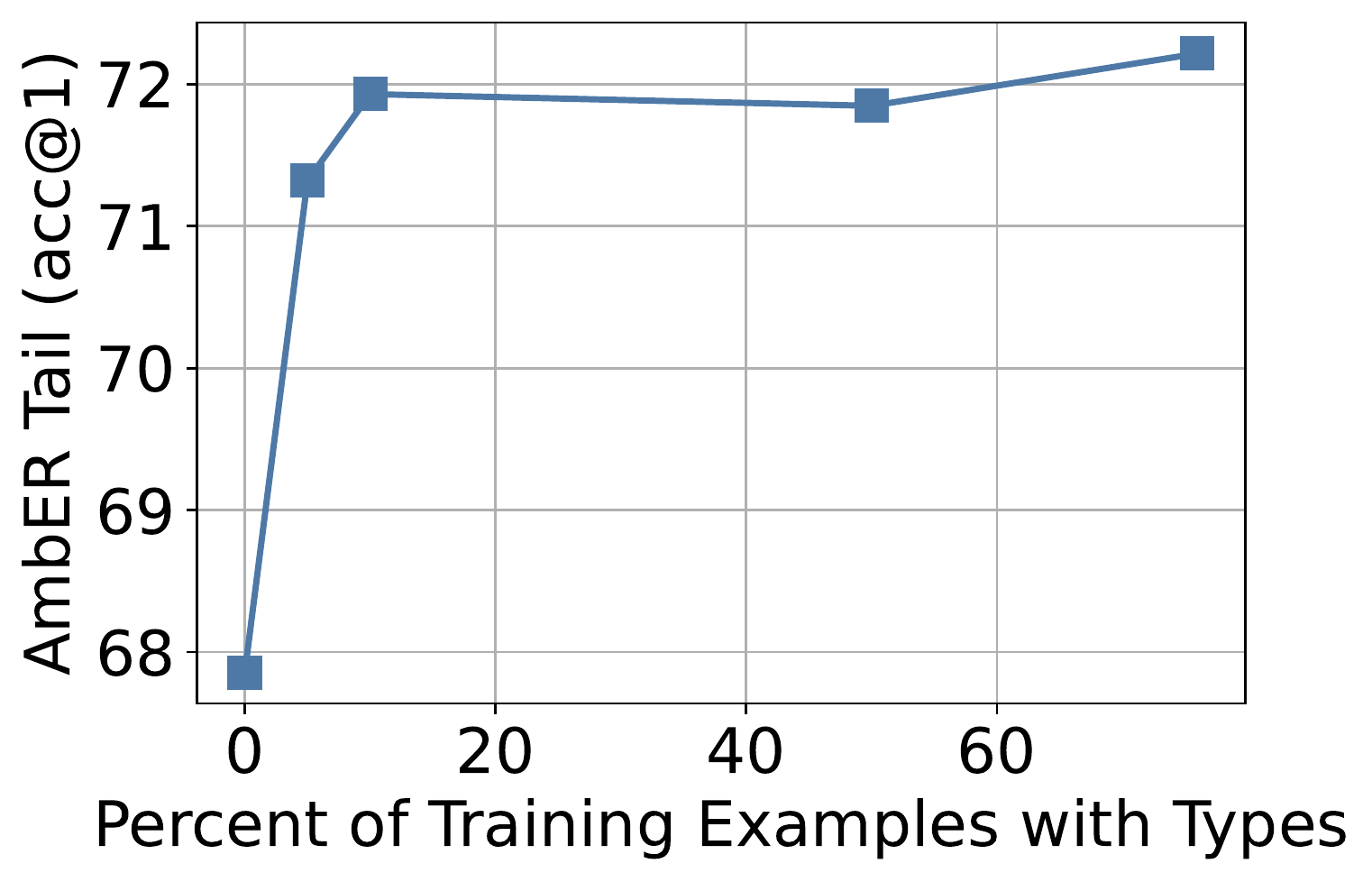}
\end{subfigure}
\begin{subfigure}[t]{0.3\textwidth}\centering
\includegraphics[height=1.2in]{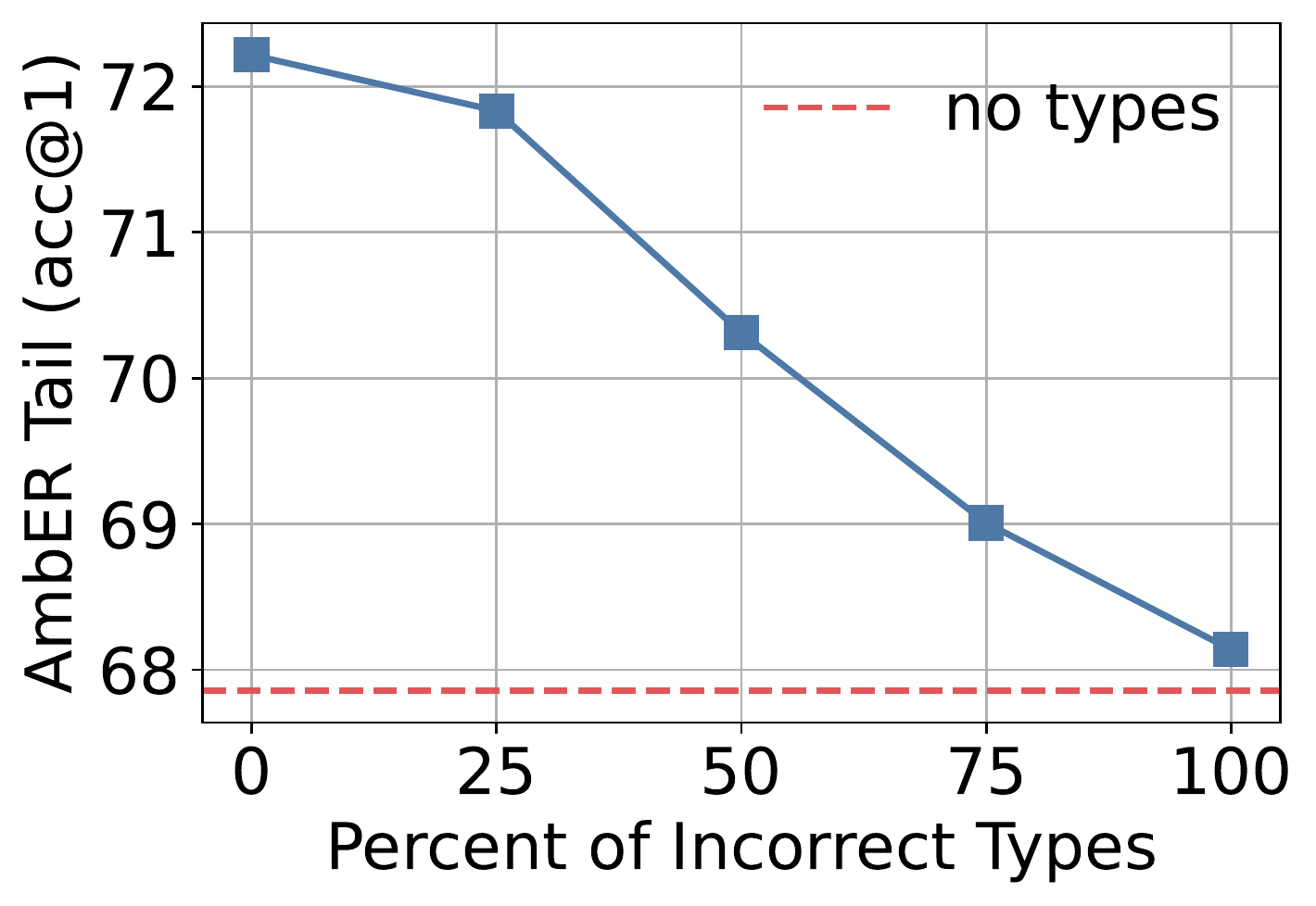}
\end{subfigure}
\begin{subfigure}[t]{0.33\textwidth}\centering
\includegraphics[height=1.2in]{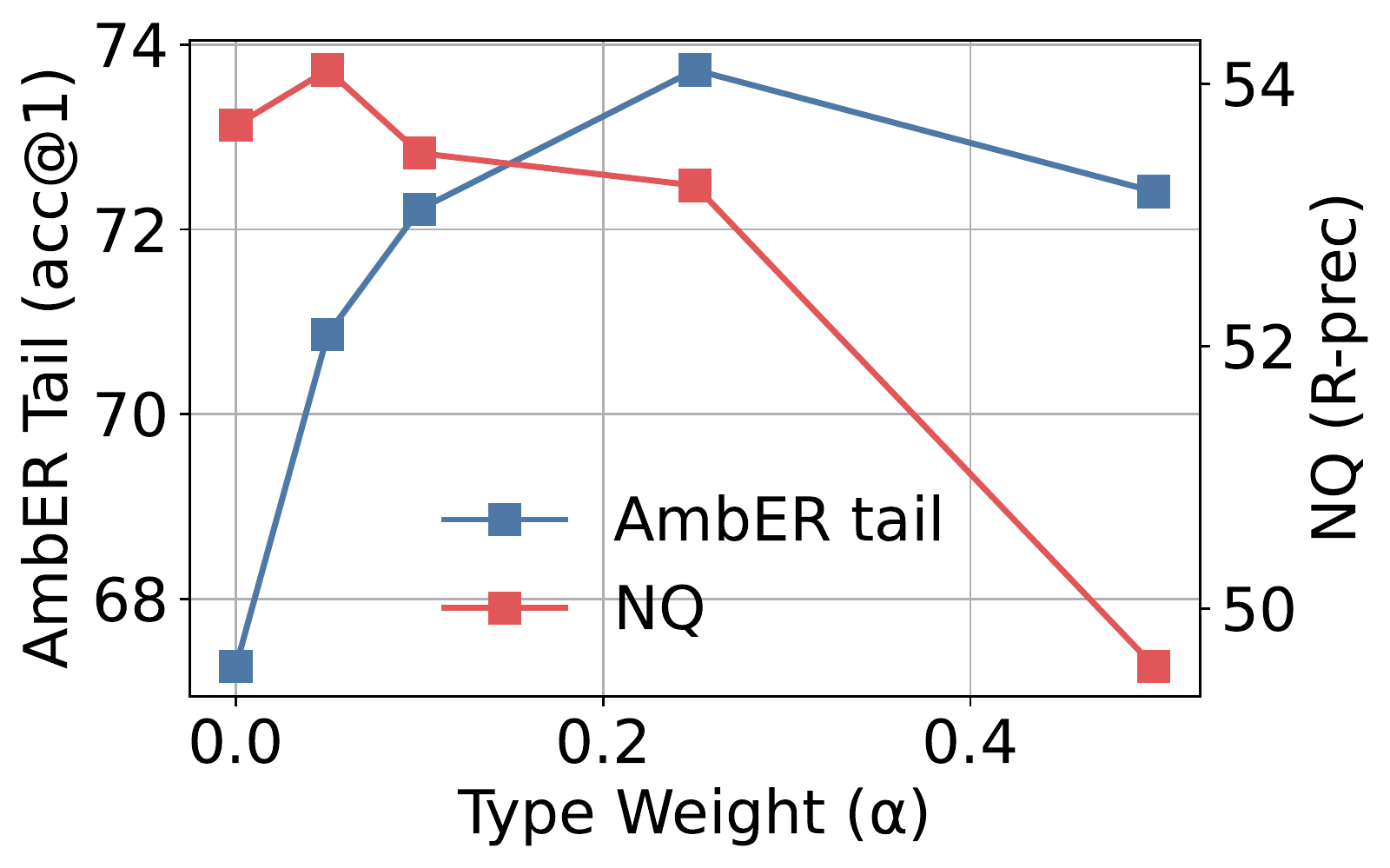}
\end{subfigure}
\caption{Robustness of \sysname\ to missing types (left) and incorrect types (middle) in the training dataset. Sensitivity of \sysname\ to the type weight $\alpha$ (right).}
\label{fig:sweep}
\end{figure*}

\paragraph{Robustness to noise}
We run two experiments to simulate incomplete and noisy type annotations.
First, we randomly remove types from a proportion of the training set.
Figure~\ref{fig:sweep} (left) shows \sysname\ achieves \yell{79}\% of the lift on AmbER tail with just \yell{5\%} type coverage.
Second, we randomly flip the types of a proportion of the training set to a type that has no type overlap with the gold type. Figure~\ref{fig:sweep} (middle) shows \sysname\ can still achieve greater than 2 points of lift over no types even when \yell{50\%} of the types are incorrect. Surprisingly, even \yell{100\%} incorrect types does not hurt performance over using no types.

\paragraph{Type weight sensitivity} Figure~\ref{fig:sweep} (right) shows \sysname's sensitivity to the type weight $\alpha$ on the AmbER tail and Natural Questions (NQ)~\cite{kwiatkowski-etal-2019-natural}, a task in KILT. We find there can be a tradeoff on some datasets: too small of an $\alpha$ is not sufficient to learn the type from the query context, whereas too large of an $\alpha$ can start to reduce overall performance. To balance this tradeoff, we set $\alpha=0.1$ in all experiments.

\paragraph{Re-ranking}
\label{ssec:rerank}
We evaluate (1) whether an inexpensive re-ranker that combines \sysname\ with sparse retrieval and popularity scores can further improve performance, and (2) whether hard negative sampling is necessary when we use a re-ranker. 
Table~\ref{tab:rerank} shows that re-ranking can improve accuracy@1 over by \yell{1.5} and \yell{0.6} points over the head and tail in AmbER, respectively. 
Without hard negative sampling, the performance of \sysname\ decreases, especially over the head. 
However, \sysname\ with the re-ranker and no hard negative sampling can still nearly match \sysname\ ($\alpha=0$)---the strong bi-encoder baseline without types---over the head and outperforms it over the tail, despite \sysname\ ($\alpha=0$) using hard negative sampling. 
This suggests that there may be alternatives to hard negative sampling, such as incorporating structured data, for achieving strong performance on some tasks.

\begin{table}[t]
\centering
\fontsize{8.4}{10.1}\selectfont 
\begin{tabular}{lcc}
\toprule
\bf{Model}         & \bf{Avg. Head} & \bf{Avg. Tail} \\
\midrule
\sysname               &  74.3         &    72.1       \\
\sysname\  ($\alpha=0$) & 71.3 & 66.3 \\
\midrule
\sysname\ + RR          &   75.8        &  72.7         \\
\sysname\ + RR (no hard negatives) &  70.4   &   70.7 \\
\bottomrule
\end{tabular}

\caption{Average head/tail accuracy@1 on AmbER when \sysname\ is combined with an inexpensive re-ranker (RR).}
\label{tab:rerank}
\end{table}

\section{Embedding Quality Analysis} 
\label{sec:embedding-quality}

\label{sec:type-eval}
We evaluate how well \sysname\ captures types through embedding visualization,  nearest neighbor type classification, and 
an entity similarity task.

\paragraph{Embedding visualization} 
\label{ssec:embed-visualization}
We use t-SNE to qualitatively evaluate how well bi-encoders cluster entity embeddings by type. 
Figure~\ref{fig:tsne} shows that \sysname\ forms tighter type clusters than BLINK for five FIGER types.\footnote{We choose popular types with low overlap in entities.} 
Types are not captured as well when the type is only present in the input and not the loss.
This suggests that our type-based loss term helps encode types in the embedding space.

\begin{table}[t]
\fontsize{8.4}{10.1}\selectfont 
\center
\begin{tabular}{llrrr}
\toprule
Dataset   & Model & Acc. & Micro F1 & Macro F1 \\
\midrule
FIGER     & BLINK & 15.8     & 40.5     & 25.1     \\
          & \bf{\sysname}  & \bf{49.0}     & \bf{72.8}     & \bf{76.6}     \\
\midrule
OntoNotes & BLINK & 21.5     & 34.2     & 42.3     \\
          & \bf{\sysname} &   \bf{38.6}     &    \bf{57.3}  & \bf{63.3}   \\
\bottomrule
\end{tabular}
\caption{Mention type classification using a nearest neighbor classifier over query embeddings.}
\label{tab:men-type}
\end{table}

\paragraph{Type classification} 
\label{ssec:nn-classification}
To better understand how well embeddings are clustered by type, we evaluate query and entity embeddings 
using KNN classification with $K=10$.\footnote{As a query or entity can have multiple types, we cast type classification as a multi-label classification problem.} We use strict accuracy, loose micro F1, and loose macro F1 metrics for evaluation~\cite{zhang-etal-2019-ernie}.  
\sysname\ outperforms BLINK on KNN classification over query embeddings on FIGER and OntoNotes, confirming that our loss encourages nearby query embeddings to share the same type (Table~\ref{tab:men-type}). Appendix~\ref{appendix:entity-type-classification} reports KNN experiments on entity embeddings, where we find \sysname\ outperforms BLINK on KNN classification of both coarse and fine types, confirming our loss also helps the entity embeddings encode types.

\paragraph{Entity similarity ranking} 
\label{ssec:entity-similarity}
To understand how well our method learns finer-grained type hierarchies, we create a novel entity similarity task inspired by word similarity tasks~\cite{schnabel-etal-2015-evaluation}. 
The goal is to rate the similarity of entity pairs, where a pair has a high score if the two entities share a fine type and a lower score otherwise. 
We assign ground truth similarity scores to 500 entity pairs that share Wikidata types\footnote{We use the "instance of" (P31), "subclass of" (P279), and "occupation" (P106) relations to extract types from Wikidata.} of varying coarseness using a weighted Jaccard similarity metric from the KGTK Semantic Similarity toolkit~\cite{ilievski2021userfriendly}\footnote{\url{https://github.com/usc-isi-i2/kgtk-similarity}}(see Appendix~\ref{appendix:entity-similarity}).

Table~\ref{tab:entity_similarity} compares the Spearman rank correlation of the inner products of BLINK and \sysname\ entity embeddings with the ground truth similarity scores, as well as two popular knowledge graph embeddings, TransE~\cite{Bordes2013TranslatingEF} and ComplEx~\cite{Trouillon2016ComplexEF} (for which we use cosine similarities between entity pairs provided by KGTK).
\sysname\ outperforms BLINK and the knowledge graph embeddings. 
This is surprising, since the knowledge graph embeddings are trained on triples which include Wikidata types, whereas \sysname\ is only trained with coarser-grained FIGER types.

\begin{table}[t]
\centering
\fontsize{8.4}{10.1}\selectfont 
\begin{tabular}{lcccc}
\toprule
 & TransE & ComplEx & BLINK & \bf{\sysname} \\
\midrule
Spearman $\rho$ & 62.4 & 63.4 & 59.4 & \bf{68.6} \\
\bottomrule
\end{tabular}
\caption{Spearman rank correlation on our proposed entity similarity task over pairs of Wikidata entities.}
\label{tab:entity_similarity}
\end{table}

\section{Discussion} 
\label{sec:discussion}

We discuss limitations of \sysname.
First, we assume a relatively coarse type system is available.
To pull together query embeddings of the same type, the type system needs to be sufficiently coarse-grained and the batch size large enough such that multiple examples in a randomly sampled batch have the same type.  
Second, our method is designed for open-domain tasks, which tend to have short queries and strong type disambiguation signals.
However, there are disambiguation signals that may be present in queries, such as the existence of a knowledge graph relation between two entities, that \sysname\ does not optimize for learning.
To address this, we are interested in incorporating other forms of structured data, including different modalities, into our model as future work.

\section{Related Work} 
\label{sec:related}

\paragraph{Entity disambiguation with types} Our work is inspired by prior work that has used types for entity disambiguation~\cite{ling-etal-2015-design,gupta-etal-2017-entity,gillick-etal-2019-learning,onoe,Chen2020ImprovingEL,bootleg}. 
Most closely related are \citet{gillick-etal-2019-learning} and \citet{gupta-etal-2017-entity}.
\citet{gillick-etal-2019-learning} train dense retrievers with Wikipedia categories as input, but do not include types in the loss function. 
On the other hand, \citet{gupta-etal-2017-entity} incorporate types through multi-task learning with type prediction, but rely on alias tables.
Generally, prior works that use types assume mention boundaries are given as input. 
Similar to our work, \citet{gupta-etal-2017-entity}, \citet{onoe}, and \citet{bootleg} show that using types can improve disambiguation of rare entities. In addition to entity disambiguation, types have also been shown to improve performance on coreference resolution~\cite{khosla-rose-2020-using} and natural language generation~\cite{dong-etal-2021-injecting}. Finally, we refer the reader to \citet{typeaware} for a study of how the type taxonomy impacts entity retrieval for keyword user queries.

\paragraph{Entity typing} A task closely related to our work is entity typing, or predicting the set of types for a mention (e.g., ~\citealp{Ling2012FineGrainedER,Gillick2014ContextDependentFE,onoe-etal-2021-modeling}). 
A key difference is that entity typing methods often learn explicit type embeddings to perform type classification, whereas \sysname\ only learns query and entity embeddings.
Entity typing methods could be used to add type labels to the training data as an alternative to \sysname's approach that uses a direct knowledge graph type mapping. 

\paragraph{Retrieval for open-domain NLP} There has been extensive work on dense retrieval for open-domain NLP tasks (e.g.~\citealp{lee-etal-2019-latent,karpukhin-etal-2020-dense,Ouz2020UniKQAUR}). 
However, most prior work has assumed unstructured text as the only input.  
As an exception, \citet{Ouz2020UniKQAUR} incorporate structured data, such as knowledge graph relations and tables, into dense retrieval by flattening the structured data into text and adding it to the retrieval index.
This approach is complementary to \sysname, which incorporates the structured data into the loss to learn better representations of the index.

\paragraph{Alternatives to bi-encoders} 
Several works have focused on improving the bi-encoder model by leveraging multiple embeddings for each query or candidate~\cite{Humeau2020PolyencodersAA,Khattab2020ColBERTEA,Luan2021SparseDA}.  
These approaches are complementary to \sysname---which maintains a single embedding for each query and candidate---and may lead to further quality improvements at some computational expense.

\section{Conclusion}

We introduce a method to train bi-encoders on unstructured text and knowledge graph types through a type-enforced contrastive loss. 
Our loss can improve retrieval of rare entities for ambiguous mentions, while maintaining strong overall performance on open-domain NLP tasks.
We hope our work inspires future work on integrating structured data into pretrained models.

\section*{Acknowledgements}

 We thank Simran Arora, Ines Chami, Neel Guha, Laurel Orr, Maya Varma, Sen Wu, and the anonymous reviewers for their helpful feedback. 
We gratefully acknowledge the support of NIH under No. U54EB020405 (Mobilize), NSF under Nos. CCF1763315 (Beyond Sparsity), CCF1563078 (Volume to Velocity), and 1937301 (RTML); ARL under No. W911NF-21-2-0251 (Interactive Human-AI Teaming); ONR under No. N000141712266 (Unifying Weak Supervision); ONR N00014-20-1-2480: Understanding and Applying Non-Euclidean Geometry in Machine Learning; N000142012275 (NEPTUNE); NXP, Xilinx, LETI-CEA, Intel, IBM, Microsoft, NEC, Toshiba, TSMC, ARM, Hitachi, BASF, Accenture, Ericsson, Qualcomm, Analog Devices, Google Cloud, Salesforce, Total, the HAI-GCP Cloud Credits for Research program,  the Stanford Data Science Initiative (SDSI), the NSF Graduate Research Fellowship under No. DGE-1656518, the Department of Defense under the National Defense Science and Engineering Graduate Fellowship Program, and members of the Stanford DAWN project: Facebook, Google, and VMWare. The U.S. Government is authorized to reproduce and distribute reprints for Governmental purposes notwithstanding any copyright notation thereon. Any opinions, findings, and conclusions or recommendations expressed in this material are those of the authors and do not necessarily reflect the views, policies, or endorsements, either expressed or implied, of NIH, ONR, or the U.S. Government.

\section*{Broader Impact}
We believe that our work has the potential to positively impact underrepresented populations. 
A key benefit of our method is improved retrieval of rare entities, which infrequently or never occur in the training dataset. 
Rare entities may not only consist of individuals from underrepresented populations, but may also be entities that are of interest to underrepresented populations (e.g., songs, locations). 
While we hope our work will have a positive impact, we also caution that our method is susceptible to biases present in standard pretrained language models and large Internet-based training datasets. 
We fine-tune our model from a BERT-base pretrained model using BLINK and KILT training datasets, which include content from Wikipedia, Reddit, trivia websites, and crowd-sourced questions and dialogue.

\bibliography{anthology,custom}
\bibliographystyle{plainnat}

\appendix
\clearpage
\section*{Appendix}
\appendix

\section{Experimental Setup Details} 
\label{appendix:experimental-setup}

\subsection{Baselines}
\label{appendix:baselines} 

We use pretrained models for all learned text-only baselines. For fair comparison, we use the KILT-E knowledge base at inference time for all models (see Section~\ref{ssec:experimental-setup} for details on the knowledge base). We include model parameter counts in Table~\ref{tab:model-parameters}. Note that DPR, BLINK (Bi-encoder), BLINK, ELQ, \sysname-type-text, and \sysname\ require an index of embeddings to be stored in addition to the model parameters for fast inference. 

\begin{table}[h]
\centering
\small
\begin{tabular}{lr}
\toprule
Model      & \# Parameters \\
\midrule
\emph{Text-only methods} \\
Alias Table  & 0 \\
TF-IDF     & 0             \\
DPR        & 220M          \\
BLINK (Bi-encoder) & 680M   \\
BLINK      & 1.0B        \\
ELQ & 680M             \\
GENRE      & 406M          \\
\midrule 
\emph{Type-aware methods} \\
Bootleg    & \yell{1.3B}           \\
GENRE-type  & 406M          \\
\sysname-type-text     & \yell{110M}         \\
\sysname\     & \yell{110M}         \\
\bottomrule
\end{tabular}
\caption{Number of model parameters.} 
\label{tab:model-parameters}
\end{table}

For Alias Table, we compute the prior probability of a mention-entity pair over the BLINK training dataset. 

For TF-IDF, DPR, and BLINK, we use the code provided in the KILT repository.\footnote{\url{https://github.com/facebookresearch/KILT}} For the BLINK cross-encoder, we use $k=10$ as the number of retrieved entities passed to the cross-encoder, following the recommended setting in \citet{wu-etal-2020-scalable}. BLINK uses Flair~\cite{akbik-etal-2019-flair} for mention detection when no mention boundaries are available.

For ELQ, we use the code provided in the ELQ repository.\footnote{\url{https://github.com/facebookresearch/BLINK/tree/main/elq}} We use the Wikipedia-trained ELQ model and the recommended settings for the Wikipedia model provided in the repository (threshold: -2.9). We find this outperforms the WebQSP-finetuned ELQ model on average on AmbER and KILT.

For Bootleg, we use the code provided in the Bootleg repository.\footnote{\url{https://github.com/HazyResearch/bootleg}} We use the model version from July 2021. Bootleg uses a heuristic n-gram method for mention detection when no mention boundaries are available.

For GENRE, we use the code provided in the GENRE repository.\footnote{\url{https://github.com/facebookresearch/GENRE}} We use the BLINK-trained model for experiments on AmbER (GOLD) and the KILT-trained model for experiments on AmbER and KILT. We use the default settings (beam size: 10, context length: 384 tokens).

For GENRE-type, we modify GENRE so that instead of just generating the entity name, the model must generate the entity name and type to predict an entity (e.g. ``United States country"). First, we use the FIGER types from KILT-E to generate a new set of type-enhanced titles. 
We then train models for both the AmbER and AmbER (GOLD) settings. For AmbER experiments, we fine-tune from the GENRE KILT-pretrained model for 4 epochs on the KILT dataset. We set max tokens to 8,192 and train on 16 A100s. We sweep the learning rate in \{1e-4, 3e-5, 1e-5, 1e-6\} and select the best value on the KILT dev set using the macro-average R-precision across the eight open-domain tasks (best learning rate: 1e-6). For AmbER (GOLD) experiments, we fine-tune from the GENRE BLINK-pretrained model for 4 epochs on the BLINK dataset using the same learning rate (1e-6). For both models, we run inference using a trie created over the type-enhanced titles and a maximum output length of 20 tokens. 

For \sysname-type-text, we use the type as textual input to the entity encoder and do not use types in the loss function. Specifically, we insert the types after the entity title and before the description, separated by a special separator token. We use $L_{ent}$ for training \sysname-type-text and use the same training procedure as we use for \sysname\ described in Appendix~\ref{appendix:training-procedure}. We fix the temperature to 0.05 and batch size to 4,096. We sweep the learning rate in \{1e-4, 2e-4, 3e-4\} for two epochs on the KILT training data and select the best value on the KILT dev set using the macro-average R-precision across the eight open-domain tasks (best learning rate: 2e-4). We use the same learning rate to train a model on the BLINK training data.

For all models, we report a single run. 

\subsection{Evaluation datasets}
\label{appendix:eval-datasets}
We include statistics on the evaluation datasets described in Section~\ref{ssec:experimental-setup} in Table~\ref{tab:data-stats}. We report the head/tail subsets for AmbER as defined in \citet{chen-etal-2021-evaluating}. Note we split AmbER randomly into dev (5\%) and test (95\%) splits and report results on test. We consider the open-domain tasks in KILT (fact checking, question answering, slot filling, and dialogue) 
and define the ``head" as having a gold entity that is in the top 1\% most popular entities by Wikipedia page views and the ``tail" as being in the bottom 90\% of entities by Wikipedia page views. 
We evaluate retrieval on eight datasets: FEVER~\cite{thorne-etal-2018-fever}, T-REx~\cite{elsahar-etal-2018-rex}, Zero Shot RE~\cite{levy-etal-2017-zero}, Natural Questions~\cite{kwiatkowski-etal-2019-natural}, HotPotQA~\cite{yang-etal-2018-hotpotqa},  TriviaQA~\cite{joshi-etal-2017-triviaqa}, ELI5~\cite{fan-etal-2019-eli5}, and Wizard of Wikipedia~\cite{dinan2019wizard}.

\begin{table*}[]
\centering
\fontsize{8.4}{10.1}\selectfont 
\setlength{\tabcolsep}{4pt}
\begin{tabular}{ll rrr rrr l}
\toprule

                        &                     & \multicolumn{3}{c}{Dev} & \multicolumn{3}{c}{Test} &                                    \\
\cmidrule(lr){3-5}
\cmidrule(lr){6-8}
Benchmark               & Dataset             & Total   & \# Head  & \# Tail  & Total  & \# Head   & \# Tail   & Type of Queries                    \\
\midrule

\multirow{6}{*}{AmbER} & Human FC            & 594     & 284   & 310   & 11,290 & 5,054  & 6,236  & Templated claims                   \\
                        & Non-human FC        & 1,369   & 728   & 641   & 26,017 & 13,500 & 12,517 & Templated claims                   \\
                        & Human SF            & 297     & 138   & 159   & 5,645  & 2,531  & 3,114  & Subject-relation facts             \\
                        & Non-human SF        & 684     & 355   & 329   & 13,009 & 6,759  & 6,250  & Subject-relation facts             \\
                        & Human QA            & 297     & 123   & 174   & 5,645  & 2,546  & 3,099  & Templated questions                \\
                        & Non-human QA        & 684     & 343   & 341   & 13,009 & 6,771  & 6,238  & Templated questions                \\
                        \midrule
\multirow{8}{*}{KILT}   & FEVER               & 10,444  & 6,406 & 614   & 10,100 & -      & -      & Mutated Wikipedia claims           \\
                        & T-REx               & 5,000   & 35    & 4,553 & 5,000  & -      & -      & Subject-relation facts             \\
                        & Zero Shot RE        & 3,724   & 111   & 2,974 & 4,966  & -      & -      & Subject-relation facts             \\
                        & Natural Questions   & 2,837   & 1,444 & 204   & 1,444  & -      & -      & Search engine questions            \\
                        & HotpotQA            & 5,600   & 2,115 & 797   & 5,569  & -      & -      & Crowd-sourced questions            \\
                        & TriviaQA            & 5,359   & 3,747 & 223   & 6,586  & -      & -      & Trivia questions from trivia sites \\
                        & ELI5                & 1,507   & 644   & 168   & 600    & -      & -      & Reddit questions                   \\
                        & Wizard of Wikipedia & 3,054   & 1,963 & 142   & 2,944  & -      & -      & Crowd-sourced dialogue  \\      
\bottomrule
\end{tabular}
\caption{Evaluation dataset statistics.}
\label{tab:data-stats}
\end{table*}

\subsection{Training data}
We include additional details about the training data described in Section~\ref{ssec:experimental-setup}.

\paragraph{Unstructured text} In the BLINK training data, each sentence has a single mention labeled with mention boundaries and a gold entity from a Wikipedia anchor link. 
The KILT training data is a superset of the BLINK training data, that additionally contains sentences from standard fact checking, slot filling, open domain QA, dialogue, and entity disambiguation datasets. 
With the exception of the entity disambiguation examples, the additional examples have a gold entity label, but no gold mention boundaries.

\paragraph{Knowledge graph types} We describe (1) how we assign types to entities and (2) how we assign types to queries. For both entities and queries, we use the FIGER type set~\cite{Ling2012FineGrainedER} for types (i.e., each type label must be one of 113 types in the type set); however, our method is not specific to the FIGER type set and any type set with coarse types may lead to improvements.  

\textit{Entity type assignments} We assign types to entities via a direct mapping of entities to knowledge graph types. The majority of the entities in KILT-E have a unique QID in Wikidata. For these entities, we use the ``P646" property in Wikidata to map from Wikidata to Freebase. After finding the corresponding Freebase entity, we derive the FIGER types from its Freebase types, using the map from \citet{Ling2012FineGrainedER}.

\textit{Query type assignments} We follow \citet{Ling2012FineGrainedER} to assign types to queries through distant supervision. Specifically, we assign the types of the gold entity for the query as the types of the query. Thus we use a direct mapping of entity types from a knowledge graph, rather than use a probabilistic type classifier. Note that assigning query types through distant supervision (with the gold entity types) can be a noisy assumption. For instance, consider the query \emph{``What was the outcome of the election for Arnold Schwarzenegger?''} with the gold entity Arnold Schwarzenegger. The query only implies that Schwarzenegger is a politician with the keyword ``election''. However, all types of the gold entity Arnold Schwarzenegger would be assigned to the query (e.g. ``actor'', ``body builder'', assuming the types were in the type set). As not all types associated with the gold entity may be implied by the query, this method can add noise to the query type labels.

\textit{Type statistics} We are able to assign types to \yell{73\%} of examples in the BLINK training data and \yell{76\%} of examples in the KILT training data. In the BLINK training data, the average example with types has \yell{2.1} types with a max of \yell{9} types. In KILT training data, the average example with types has \yell{2.0} types with a max of \yell{9} types. 

\subsection{Training procedure} 
\label{appendix:training-procedure}
We describe the training procedure for \sysname. We tie the query and entity encoders (i.e. use a single encoder) and initialize from a BERT-base pretrained model~\cite{devlin-etal-2019-bert}. 
Following BLINK's protocol~\cite{wu-etal-2020-scalable}, we set the maximum context length to 32 tokens and the maximum entity description length to 128 tokens.
We set the batch size to \yell{4,096} and use the AdamW optimizer~\cite{Loshchilov2019DecoupledWD} and decay the learning rate by 50\% every epoch. 

We use balanced hard negative sampling, following \citet{botha-etal-2020-entity}. Specifically, we only allow ten negative examples of an entity for each positive example in the training dataset. For all models of \sysname, we train the first epoch using local in-batch negatives, and we mine for hard negatives at the end of every epoch. Starting at the second epoch, we train with both in-batch and hard negatives. 

From results on preliminary experiments, we fix the temperature to 0.05 and the type weight $\alpha$ to 0.1. 
We then conduct a grid search for the initial learning rate by training for two epochs on the KILT training data and selecting the best value on the KILT dev set using the macro-average R-precision across the eight open-domain tasks. We sweep the initial learning rate in $\{$1e-4, 2e-4, 3e-4$\}$ (best learning rate: 3e-4).

We use the same hyperparameter configuration for training on both the BLINK training data and the KILT training data. For models trained on BLINK data and KILT data, we train for 4 epochs using 16 A100 GPUs (approximately \yell{2.2} hours/epoch for BLINK training data, \yell{2.6} hours/epoch for KILT training data, including sampling for hard negatives).

\subsection{Re-ranking details}
\label{appendix:rerank}

While our standard configuration of \sysname\ does not use a re-ranker, we explore using an inexpensive re-ranker on top of \sysname. The re-ranker consists of two steps: first, it linearly combines the top-$K$ entity scores from the bi-encoder with the top-$K$ entity scores of a sparse retriever using a tunable weight $\lambda$. Second, it linearly combines these scores with their corresponding global entity popularity (e.g. Wikipedia page views) using a tunable weight $\kappa$. We normalize scores before linearly combining at each step.

More formally, let $E$ be the union of the set of retrieved entities from the bi-encoder and the sparse retriever. Then for an entity $e \in E$, where $s_e$ indicates the score from the sparse retriever, $d_e$ indicates the score from the dense retriever, and $p_e$ indicates the popularity score, we compute the re-ranked score $f_e$ as follows:  
\begin{align*} 
h_e = \lambda s_e  + d_e \\
f_e = \kappa p_e + h_e 
\end{align*} 

We use the baseline TF-IDF retriever for the sparse retriever (see Appendix~\ref{appendix:baselines} for details). Like \citet{chen-etal-2021-evaluating}, we use the monthly Wikipedia page views (from October 2019) as the measure of global entity popularity.  Note that tuning these weights does not require re-training or re-running the bi-encoder evaluation.

For the experiments with the re-ranker, we tune $\lambda$ and $\kappa$ on each of the 6 dev sets for AmbER by first selecting $\lambda$ that performs best on the linear combination of the bi-encoder and sparse retriever scores, and then fixing $\lambda$ and tuning $\kappa$. For both $\lambda$ and $\kappa$, we sweep in $\{0.0, 0.25, 0.5, 0.75, 1.0, 1.25, 1.5, 1.75, 2.0\}$.

\section{Extended Retrieval Results}

\subsection{AmbER results} 
\label{appendix:extended_amber}

\begin{table*}[t]
\centering
\fontsize{8.4}{10.1}\selectfont 
\setlength{\tabcolsep}{3.25pt}
\begin{tabularx}{\textwidth}{l rr rr rr rr rr rr rr}
\toprule
          & \multicolumn{4}{c}{\bf{Fact Checking}}   & \multicolumn{4}{c}{\bf{Slot Filling}}        & \multicolumn{4}{c}{\bf{Question Answering}} &    &      \\
          & \multicolumn{2}{c}{H}     & \multicolumn{2}{c}{N}   & \multicolumn{2}{c}{H}     & \multicolumn{2}{c}{N} & \multicolumn{2}{c}{H}     & \multicolumn{2}{c}{N} & \multicolumn{2}{c}{\bf{Average}} \\
          \cmidrule(lr){2-3}
          \cmidrule(lr){4-5}
          \cmidrule(lr){6-7}
          \cmidrule(lr){8-9}
          \cmidrule(lr){10-11}
          \cmidrule(lr){12-13}
          \cmidrule(lr){14-15}
          \bf{Model} & Head & Tail & Head & Tail & Head & Tail & Head & Tail & Head               & Tail & Head & Tail & Head    & Tail \\
\midrule  
BLINK (Bi-encoder) + Flair         & 56.4 & 52.0 & 24.8 & 10.5 & 76.8 & 55.7 & 30.7 & 13.5 & 78.3 & 55.7 & 67.3 & 33.8 & 55.7 & 36.9 \\
\sysname\ ($\alpha=0$, BLINK data) + Flair & 45.6 & 49.7 & 4.6  & 2.9  & 77.4 & 58.1 & 48.4 & 30.4 & 78.0 & 58.0 & 63.4 & 39.9 & 52.9 & 39.8 \\
\sysname\ ($\alpha=0$, KILT data) + Flair  & 44.7 & 44.9 & 7.5  & 7.4  & 80.0 & 84.0 & 74.5 & 73.6 & 83.4 & 70.2 & 77.6 & 56.5 & 61.3 & 56.1 \\
\sysname\ ($\alpha=0$, KILT data)  & 77.6 & 61.9 & 41.4 & 39.1 & 70.9 & 87.1 & 83.2 & 85.9 & 72.5 & 66.3 & 82.2 & 57.7 & 71.3 & 66.3  \\
\bottomrule
\end{tabularx}
\caption{Retrieval accuracy@1 on AmbER (H for human, N for non-human subsets). Impact of the training data on bi-encoder performance.}
\label{tab:amber-baselines}
\end{table*}

\begin{table*}[t]
\centering
\fontsize{8.4}{10.1}\selectfont 
\setlength{\tabcolsep}{5pt}
\begin{tabularx}{\textwidth}{l rr rr rr rr rr rr rr}
\toprule
          & \multicolumn{4}{c}{\bf{Fact Checking}}   & \multicolumn{4}{c}{\bf{Slot Filling}}        & \multicolumn{4}{c}{\bf{Question Answering}} &    &      \\
          & \multicolumn{2}{c}{H}     & \multicolumn{2}{c}{N}   & \multicolumn{2}{c}{H}     & \multicolumn{2}{c}{N} & \multicolumn{2}{c}{H}     & \multicolumn{2}{c}{N} & \multicolumn{2}{c}{\bf{Average}} \\
          \cmidrule(lr){2-3}
          \cmidrule(lr){4-5}
          \cmidrule(lr){6-7}
          \cmidrule(lr){8-9}
          \cmidrule(lr){10-11}
          \cmidrule(lr){12-13}
          \cmidrule(lr){14-15}
          \bf{Model} & Head & Tail & Head & Tail & Head & Tail & Head & Tail & Head               & Tail & Head & Tail & Head    & Tail \\
\midrule  
TF-IDF             & 76.4 & 76.1 & 60.9 & 60.6 & 80.4 & 82.9 & 52.6 & 50.0 & 78.1 & 82.3 & 58.9 & 54.2 & 67.9 & 67.7 \\
DPR                & 47.9 & 27.9 & 72.6 & 43.2 & 34.0 & 14.0 & 74.3 & 43.6 & 46.0 & 22.2 & 77.5 & 45.4 & 58.7 & 32.7 \\
BLINK (Bi-encoder) & 89.5 & 90.1 & 81.5 & 71.6 & 94.5 & 95.9 & 48.9 & 41.2 & 94.9 & 95.8 & 90.9 & 86.3 & 83.4 & 80.1 \\
BLINK              & 91.1 & 85.8 & 83.9 & 76.3 & 94.1 & 95.2 & 49.3 & 41.5 & 94.9 & 95.8 & 91.2 & 86.6 & 84.1 & 80.2 \\
ELQ                & 78.4 & 61.1 & 66.8 & 37.2 & 74.5 & 44.1 & 59.7 & 27.1 & 77.5 & 47.2 & 62.1 & 30.7 & 69.8 & 41.2 \\
GENRE      & 78.0 & 67.9 & 82.8 & 77.4 & 86.9 & 92.5 & 90.7 & 90.8 & 83.7 & 83.7 & 87.4 & 82.7 & 84.9 & 82.5 \\
\midrule
Bootleg$^\dagger$  & 98.3 & 97.6 & 69.9 & 65.7 & 96.5 & 93.6 & 66.8 & 56.2 & 97.1 & 96.7 & 74.8 & 76.3 & 83.9 & 81.0 \\
GENRE-type & 71.2 & 80.0 & 76.4 & 77.7 & 73.6 & 92.7 & 91.0 & 92.2 & 83.0 & 90.5 & 91.3 & 91.4 & 81.1 & 87.4 \\
\sysname-type-text & 90.9 & 83.7 & 84.5 & 77.9 & 89.2 & 95.9 & 96.2 & 98.2 & 86.0 & 88.2 & 95.1 & 92.9 & 90.3 & 89.5 \\
\bf{\sysname}    &  95.0 & 93.8 & 79.9 & 80.3 & 91.3 & 96.8 & 96.4 & 98.3 & 91.6 & 93.8 & 95.8 & 95.7 & 91.7 & 93.1 \\
\bottomrule
\end{tabularx}
\caption{Retrieval accuracy@10 on AmbER (H for human, N for non-human subsets). $^\dagger$Models with an alias table.}
\label{tab:amber-results-top10}
\end{table*}

\begin{table}[t]
\centering
\fontsize{8.4}{10.1}\selectfont 
\setlength{\tabcolsep}{4pt}
\begin{tabular}{l rr rr rr rr}
\toprule
          & \multicolumn{2}{c}{\bf{FC}}    & \multicolumn{2}{c}{\bf{SF}}      & \multicolumn{2}{c}{\bf{QA}} &         \\
            \cmidrule(lr){2-3} \cmidrule(lr){4-5} \cmidrule(lr){6-7} 
 \bf{Model}          & \multicolumn{1}{c}{H}    & \multicolumn{1}{c}{N}    & \multicolumn{1}{c}{H}    & \multicolumn{1}{c}{N}      &\multicolumn{1}{c}{H}    & \multicolumn{1}{c}{N}      & \multicolumn{1}{c}{\bf{Avg.}} \\
\midrule
TF-IDF             & 1.0  & 0.6 & 2.5  & 2.5  & 2.5  & 2.5  & 1.9  \\
DPR                & 0.2  & 3.8 & 1.2  & 10.7 & 2.3  & 12.2 & 5.1  \\
BLINK (Bi-enc) & 9.4  & 0.7 & 36.1 & 6.4  & 35.9 & 20.5 & 18.2 \\
BLINK              & 5.4  & 0.0 & 17.6 & 8.6  & 27.7 & 29.7 & 14.8 \\
ELQ                & 3.9  & 0.0 & 24.7 & 12.4 & 29.6 & 16.2 & 14.5 \\
GENRE              & 4.3  & 1.0 & 28.3 & 39.2 & 10.9 & 13.9 & 16.3 \\
\midrule
Bootleg            & 3.0  & 0.0 & 26.7 & 15.5 & 31.6 & 27.8 & 17.4 \\
GENRE-type & 3.3 &	7.4	& 17.2 &	50.9 &	15.8 &	28.6 &	20.5 \\
\sysname-type-text & 17.3 &	2.1 &	60.0 &	69.1 &	40.9 &	44.8 & 39.0 \\ 
\bf{\sysname}            & 40.0 & 4.2 & 65.6 & 74.3 & 53.6 & 52.6 & 48.4 \\
\bottomrule 
\end{tabular}
\caption{Consistency results on AmbER for top-1. The consistency is the fraction of mentions where all queries for a mention are correct.} 
\label{tab:consistency}
\end{table}

We extend the results on AmbER included in Section~\ref{sec:experiments}. First, we perform experiments to better understand the strong performance of the baseline \sysname\ ($\alpha=0$), which removes the type-based loss term. We primarily attribute the strong performance of \sysname\ ($\alpha=0$) relative to the BLINK (Bi-encoder) to the training data and perform baseline ablations in  Table~\ref{tab:amber-baselines}. We see that BLINK (Bi-encoder) and \sysname\ ($\alpha=0$) perform similarly when both are trained on BLINK data, which consists of Wikipedia entity disambiguation data. Training on the KILT data, which additionally includes multiple open-domain tasks, leads to significant lift. Removing the mention detector, Flair, leads to additional lift. Note that \sysname\ ($\alpha=0$) can retrieve entities without mention detection since the KILT training data includes open-domain tasks which do not have mention boundaries.

Second, we include results for top-10 retrieval accuracy (accuracy@10) on AmbER to understand the retrieval performance at larger $K$ (Table~\ref{tab:amber-results-top10}). We find that \sysname\ continues to outperform baselines on average. 

Finally, we report results for the consistency metric introduced in \citet{chen-etal-2021-evaluating} for top-1 retrieval in Table~\ref{tab:consistency}. This metric measures the proportion of mentions where all queries for the mention are correct. In particular, \citet{chen-etal-2021-evaluating} found that retrievers have a tendency to ``collapse" all predictions for a mention to the most popular entity for the mention, which would result in a low consistency value. We find that \sysname\ outperforms all models on this metric.

\subsection{KILT results} 
\label{appendix:extended_kilt}

\begin{table*}[t]
\centering
\fontsize{8.4}{10.1}\selectfont 
\setlength{\tabcolsep}{9pt}
\begin{tabular}{lcccccccccc}
\toprule
 & \multicolumn{1}{c}{\bf{Fact Check.}}   & \multicolumn{2}{c}{\bf{Slot Filling}}        & \multicolumn{4}{c}{\bf{Question Answering}} &  \multicolumn{1}{c}{\bf{Dial.}}       \\
  \cmidrule(lr){2-2} \cmidrule(lr){3-4} \cmidrule(lr){5-8} \cmidrule(lr){9-9}
 & \multicolumn{1}{c}{FEV}  & \multicolumn{1}{c}{T-REx} & \multicolumn{1}{c}{zsRE}   & \multicolumn{1}{c}{NQ} & \multicolumn{1}{c}{HoPo}  & \multicolumn{1}{c}{TQA}  & \multicolumn{1}{c}{ELI5}  & \multicolumn{1}{c}{WoW} & \bf{Avg} \\
\midrule
TF-IDF     & 48.4 & 57.4 & 72.8 & 20.1 & 43.4 & 27.8 & 4.6  & 38.8 & 39.2\\
DPR        & 57.0 & 14.9 & 44.3 & 54.5 & 25.5 & 46.2 & 16.1 & 26.9 & 35.7\\
BLINK (Bi-encoder) & 64.4 & 59.4 & 84.3 & 35.1 & 43.1 & 61.6 & 11.3 & 26.0 & 48.2 \\
BLINK     & 67.6 & 61.0 & 87.4 & 33.5 & 47.9 & 65.9 & 9.7  & 26.5 & 49.9 \\
ELQ        & 65.1 & 71.2 & 95.0 & 42.4 & 45.9 & 67.7 & 9.2  & 26.8 & 52.9 \\
GENRE      & 85.0 & 80.5 & 95.1 & 61.4 & 51.9 & 71.4 & 13.6 & 56.5 & 64.4 \\
\midrule 
Bootleg$^\dagger$   & 62.3 & 69.4 & 81.8 & 34.5 & 43.6 & 53.1 & 9.7  & 28.2 & 47.8 \\
GENRE-type & 55.3 &	71.9 &	80.6 &	54.5 &	37.2 &	53.6 &	11.5 &	44.5 &	51.1 \\ 
\sysname-type-text & 87.3 &	82.2 & 95.1 & 62.5	& 51.2 &	70.8 &	16.9 &	51.0 &	64.6 \\ 
\bf{\sysname} & 85.8 & 82.0 & 95.2 & 62.4 & 52.7 & 71.5 & 16.7 & 51.8 & 64.8 \\
\bottomrule
\end{tabular}
\caption{R-precision on KILT open-domain tasks (dev data). (Top) text-only methods and (bottom) type-aware methods. $^\dagger$Models with an alias table.}
\label{tab:kilt-dev-rprecision-results}
\end{table*}

\begin{table*}[t]
\centering
\fontsize{8.4}{10.1}\selectfont 
\setlength{\tabcolsep}{10pt}
\begin{tabular}{lccccccccc}
\toprule
 & \multicolumn{1}{c}{\bf{Fact Check.}}   & \multicolumn{2}{c}{\bf{Slot Filling}}        & \multicolumn{4}{c}{\bf{Question Answering}} &  \multicolumn{1}{c}{\bf{Dial.}}       \\
   \cmidrule(lr){2-2} \cmidrule(lr){3-4} \cmidrule(lr){5-8} \cmidrule(lr){9-9}
 & \multicolumn{1}{c}{FEV} & \multicolumn{1}{c}{T-REx} & \multicolumn{1}{c}{zsRE}  & \multicolumn{1}{c}{NQ}    & \multicolumn{1}{c}{HoPo}  & \multicolumn{1}{c}{TQA}   & \multicolumn{1}{c}{ELI5}  & \multicolumn{1}{c}{WoW}   & \bf{Avg.} \\
\midrule
TF-IDF         & -    & -    & -    & -    & -    & -    & -    & -  & - \\
DPR            & 74.3 & 17.0 & 39.2 & 65.5 & 10.4 & 57.0 & 26.9 & 51.2 & 42.7 \\
Multi-task DPR & 87.5 & 83.9 & 93.1 & 68.2 & 28.4 & 68.3 & 27.5 & 67.1 & 65.5 \\
BLINK    & -    & -    & -    & -    & -    & -    & -    & -   & - \\
GENRE          & 88.2 & 85.3 & 97.8 & 61.4 & 34.0 & 75.1 & 25.5 & 77.7 & 68.1\\
KGI  & 85.0 & 83.1 & 99.2 & 70.2 & \multicolumn{1}{c}{-} & 63.5 & \multicolumn{1}{c}{-} & 78.5 & \multicolumn{1}{c}{-}\\
Re2G & 92.5 & 89.0 & \multicolumn{1}{c}{-} & 76.6 & \multicolumn{1}{c}{-} & 74.2 & - & 80.0 & \multicolumn{1}{c}{-} \\
\bf{\sysname}  & 88.6 & 89.4 & 98.7 & 64.9 & 35.5 & 69.2 & 28.2 & 69.1 & 67.9 \\
\bottomrule
\end{tabular}
\caption{Recall@5 on KILT open-domain tasks (test data). We report numbers from \citet{kilt} and the KILT leaderboard where available.}
\label{tab:kilt-test-recall-results}
\end{table*}

\begin{table*}[t]
\centering
\fontsize{8.4}{10.1}\selectfont 
\setlength{\tabcolsep}{9pt}
\begin{tabular}{lcccccccccc}
\toprule
 & \multicolumn{1}{c}{\bf{Fact Check.}}   & \multicolumn{2}{c}{\bf{Slot Filling}}        & \multicolumn{4}{c}{\bf{Question Answering}} &  \multicolumn{1}{c}{\bf{Dial.}}       \\
   \cmidrule(lr){2-2} \cmidrule(lr){3-4} \cmidrule(lr){5-8} \cmidrule(lr){9-9}
 & \multicolumn{1}{c}{FEV}  & \multicolumn{1}{c}{T-REx} & \multicolumn{1}{c}{zsRE}   & \multicolumn{1}{c}{NQ} & \multicolumn{1}{c}{HoPo}  & \multicolumn{1}{c}{TQA}  & \multicolumn{1}{c}{ELI5}  & \multicolumn{1}{c}{WoW} & \bf{Avg.} \\
\midrule
TF-IDF     & 71.8 & 73.0 & 88.6 & 32.6 & 29.2 & 41.0 & 9.7  & 56.5 & 50.3 \\
DPR        & 76.0 & 22.3 & 59.2 & 63.9 & 11.1 & 57.4 & 31.0 & 52.7 & 46.7 \\
BLINK (Bi-encoder) & 80.0 & 68.1 & 88.4 & 40.8 & 24.3 & 63.5 & 19.4 & 40.9 & 53.2\\
BLINK     & 82.9 & 69.6 & 89.6 & 43.7 & 27.4 & 66.9 & 22.3 & 44.6 & 55.9 \\
ELQ        & 79.5 & 69.9 & 95.2 & 36.1 & 23.7 & 62.4 & 9.5  & 47.7 & 53.0\\
GENRE      & 89.0 & 85.3 & 97.3 & 58.5 & 34.7 & 75.7 & 20.5 & 75.0 & 67.0 \\
\midrule 
Bootleg$^\dagger$   & 81.0 & 74.3 & 85.6 & 37.2 & 26.3 & 69.4 & 14.0 & 49.3 & 54.6\\
GENRE-type & 66.9 & 80.1 & 89.7 & 54.4 & 23.4 & 58.4 & 18.5 & 62.4 & 56.7 \\ 
\sysname-type-text & 90.6& 89.1 &	98.0 &	63.4	& 34.1 &	71.3 &	25.9 &	64.6 &	67.1 \\ 
\bf{\sysname}     & 89.3 & 88.8 & 98.3 & 63.1 & 34.2 & 70.0 & 25.6 & 64.8 & 66.8 \\
\bottomrule
\end{tabular}
\caption{Recall@5 on KILT open-domain tasks (dev data). (Top) text-only methods and (bottom) type-aware methods. $^\dagger$Models with an alias table.}
\label{tab:kilt-dev-recall-results}
\end{table*}

We include R-precision results on the KILT dev sets for the tasks and baselines in the main paper in Table~\ref{tab:kilt-dev-rprecision-results}. As with the AmbER experiments, we use the KILT-E knowledge base for inference for all models. We see that GENRE, \sysname-type-text, and \sysname\ outperform the other baselines across the tasks, and perform comparably overall to each other. Recall that GENRE, GENRE-type, \sysname-type-text, and \sysname\ were trained on KILT training data which includes open-domain tasks. BLINK, ELQ, and Bootleg were trained on Wikipedia entity disambiguation data and DPR was trained on question answering data. GENRE-type performs substantially worse than GENRE overall, suggesting that incorporating types in the entity name degrades overall retrieval performance.

We also report results on the KILT test and dev sets for recall@5. In addition to R-precision, recall@5 is reported on the KILT leaderboard and measures the proportion of gold entities for a query\footnote{The KILT benchmark supports multiple gold entities for a query.} that occur in the top-5 ranked entities. If there is a single gold entity, this is equivalent to accuracy@5. We find similar trends as seen with R-precision: \sysname, \sysname-type-text, and GENRE continue to have strong performance and outperform other baselines (Table~\ref{tab:kilt-test-recall-results} and Table~\ref{tab:kilt-dev-recall-results}).

\subsection{Impact of batch size}
\begin{figure}[t]
\centering
\includegraphics[width=0.35\textwidth]{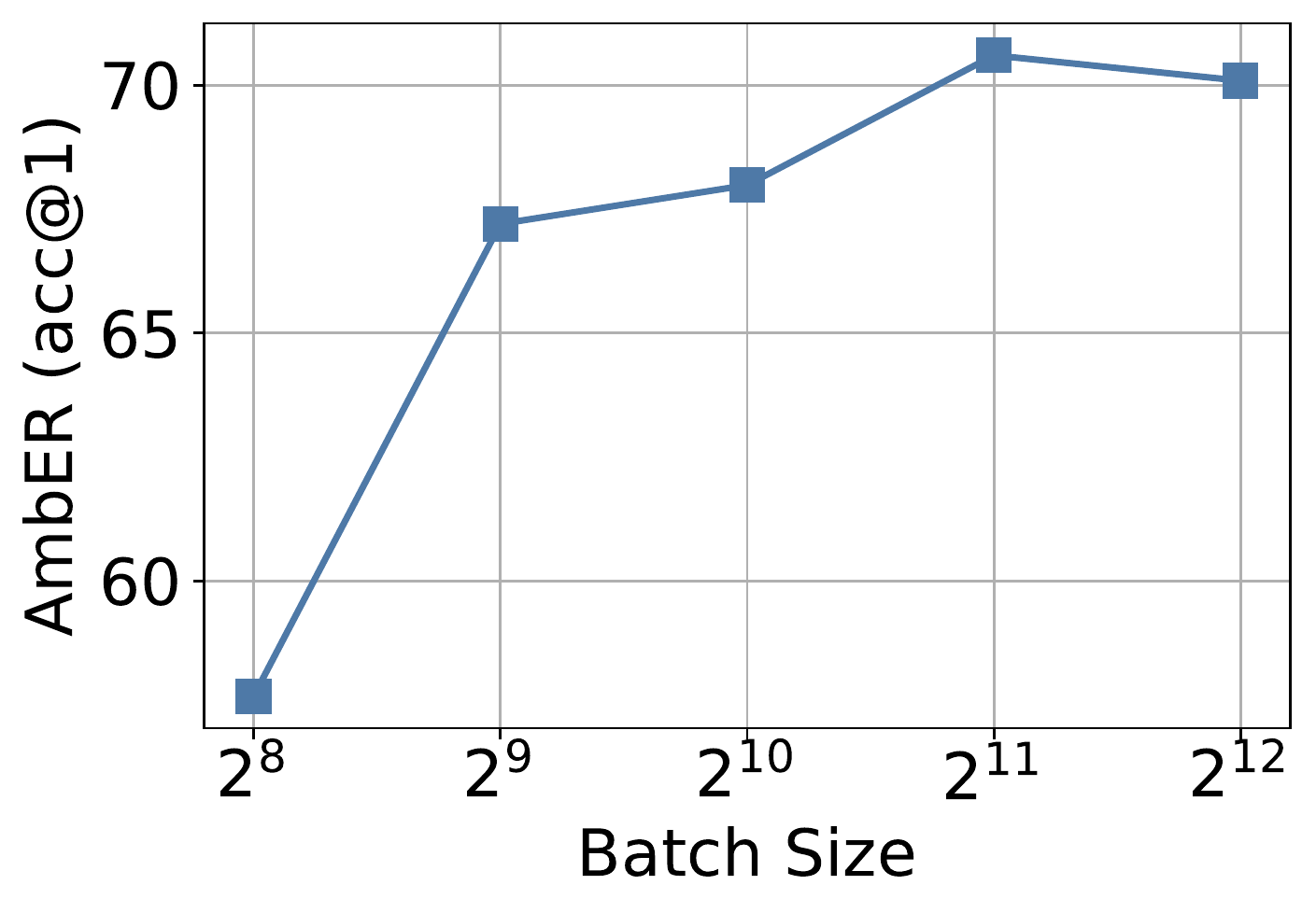}
\caption{Accuracy@1 on AmbER for varying batch sizes.}
\label{fig:batch_size}
\end{figure}

We study the impact of the batch size on \sysname\ by training on a 1M random sample of KILT training data for two epochs for batch sizes in \{256, 512, 1024, 2048, 4096\}. We hold all other hyperparameters constant. As we add $n$ hard negative samples to the batch in the second epoch, the batch size in terms of the number of queries is reduced by a factor of $n+1$. Concretely, if the base batch size is 4,096 examples and we use three hard negatives per query, each batch in the first epoch has 4,096 queries, while each batch in the second epoch has 1,024 queries. We define the batch size in terms of the number of queries in the first epoch. In Figure~\ref{fig:batch_size}, we see that generally increasing the batch size improves the average accuracy@1 on AmbER (averaged over head and tail examples). However, we caution that this study is performed with all other hyperparameters held constant and a more optimal hyperparameter configuration may exist at different batch sizes. We use a batch size of 4,096 for all experiments in the main paper.

\subsection{Type equivalence}
\label{appendix:type-equivalence}
We experiment with three type equivalence measures: (1) \emph{Any-types}: two entities have equivalent types if any types overlap, (2) \emph{All-types}: two entities have equivalent types if all types overlap, and (3) \emph{gt50-types}: two entities have equivalent types if at least 50\% of the types overlap. If the entities have an unequal number of types, then we take 50\% of the greater number of types. An example of (3) is a query A with the types [“musician”, “person”] would be considered as having equal types to query B with the types [“musician”, “person”, “author”], since more than 50\% of types of query B overlap with query A. 

Our main experiments currently use approach (3), which is intuitively a softer equivalence than (2). However, interestingly we find (2) and (3) can have very similar performance, and both greatly outperform (1). We report the average top-1 accuracy results of training the three methods for 2 epochs on a 1M random sample of KILT in Table~\ref{tab:type-equivalence}. 

\begin{table}[t]
\fontsize{8.4}{10.1}\selectfont 
\center
\begin{tabular}{lrr}
\toprule
& Avg. Head   & Avg. Tail   \\
\midrule
Any-types  & 67.9 & 63.0  \\
All-types  & 71.0 & 69.8 \\
gt50-types & 71.0 & 69.0 \\
\bottomrule
\end{tabular}
\caption{Top-1 accuracy on AmbER for different type equivalence measures.}
\label{tab:type-equivalence}
\end{table}

\section{Extended Embedding Quality Analysis}
\label{appendix:type-classification}

\subsection{Nearest neighbor mention type classification} We include additional details on the datasets used for mention type classification (experiments in Section~\ref{ssec:nn-classification}). The FIGER test set has \yell{563} examples and uses the 113 FIGER type set~\cite{Ling2012FineGrainedER}. We use the subset of the OntoNotes test set from \citet{shimaoka-etal-2017-neural} that removes pronominal mentions. We further remove examples that map to the ``other" type, resulting in a final OntoNotes test set with \yell{3,066} examples. The classifier uses \yell{50} types from the OntoNotes type set~\cite{Gillick2014ContextDependentFE} across the sampled training set and the final test set. While the training sets use distant supervision to label mentions with types over Wikipedia and news reports, respectively, both test sets consist of manually annotated mentions in news reports.  

\subsection{Nearest neighbor entity type classification}
\label{appendix:entity-type-classification}
We include the setup and results for the entity type classification task from Section~\ref{ssec:nn-classification}. 
We create two datasets for entity type classification using the \kbname\ knowledge base: Coarse-types and Fine-types. 
We use the seven coarse types in the FIGER type system as the coarse types and take the other types as fine types.
We create the Coarse-types dataset by sampling without replacement \yell{3,000} entities that correspond to the seven coarse FIGER types: ``location", ``person", ``organization", ``product", ``art", ``event", and ``building". We divide the sampled entities into training and test sets for a total of \yell{16,781} training examples and \yell{4,195} test examples. Similarly, we create the Fine-types dataset by sampling without replacement 300 entities that correspond to the FIGER fine types. 
We discard fine types that do not have at least 300 entities, leaving 100 fine types. 
We then divide the sampled entities into training and test sets for a total of \yell{23,884} training examples and \yell{5,968} test examples.

Table~\ref{tab:ent-type} reports the results for entity type classification. We find that \sysname\ outperforms BLINK, suggesting that the type-enforced loss helps cluster entities by type in the embedding space. 

\begin{table}[t]
\fontsize{8.4}{10.1}\selectfont 
\center
\begin{tabular}{llrrr}
\toprule
Dataset      & Model & Acc. & Micro F1 & Macro F1 \\
\midrule
Coarse-types & BLINK & 81.1     & 89.0     & 84.1     \\
            & \bf{\sysname} & \bf{92.7}    & \bf{95.9}  & \bf{95.9}      \\
\midrule
Fine-types   & BLINK & 71.6     & 82.0     & 77.5     \\
              & \bf{\sysname} &  \bf{76.6}    &  \bf{86.8}  &  \bf{84.0}
             \\
\bottomrule
\end{tabular}
\caption{Entity type classification using a nearest neighbor classifier over entity embeddings.}
\label{tab:ent-type}
\end{table}

\subsection{Entity similarity task} 
\label{appendix:entity-similarity} 

We describe how we construct the dataset for the entity similarity task. We first find the closure of all Wikidata types assigned to each entity in the KILT-E knowledge base. We then bucket Wikidata types by the frequency with which they occur in the KILT-E knowledge base (using five buckets). To include types of varying frequencies, we randomly sample 10 Wikidata types from each bucket (50 types total). Finally, we sample 10 pairs of entities for each type for a total of 500 entity pairs. 

To assign ``ground-truth" similarity values to each entity pair, we submit the entity pairs to the KGTK Semantic Similarity toolkit web API.\footnote{\url{https://github.com/usc-isi-i2/kgtk-similarity}} We use the Jaccard similarity metric returned by the toolkit as the ground-truth similarity. This metric assigns larger values if the types shared by two entities are more specific (i.e. fine-grained). As ground truth values are assigned automatically, there is some noise in the dataset. However, we observe that the trends on the entity similarity task generally follow the trends seen on the other embedding quality analysis tasks.

\end{document}